\documentclass[11pt,draftcls,onecolumn]{IEEEtran}
\usepackage{amsfonts}
\usepackage{mathrsfs}
\usepackage{amsfonts}
\usepackage{amssymb}
\usepackage{graphicx}
\usepackage{amsmath}
\usepackage{array}
\usepackage{cases}

\usepackage{cite,graphicx,amsmath,amssymb,color}
\usepackage{subfigure}

\newtheorem{Thm}{Theorem}
\newtheorem{Lem}{Lemma}

\newtheorem{Def}{Definition}
\newtheorem{Exam}{Example}
\newtheorem{Alg}{Algorithm}
\newtheorem{Prob}{Problem}
\newtheorem{Rem}{Remark}

\newtheorem{Asump}{Assumption}

\usepackage{multirow}

\begin{document}

\title{Queue-Aware Dynamic Clustering and Power  Allocation for Network MIMO Systems via Distributive  Stochastic Learning}
\author{\authorblockN{\small{Ying Cui, Qingqing Huang, Vincent K.~N.~Lau, \IEEEmembership{Senior~Member,~IEEE}}}\\
\authorblockA{\small{ECE Department, Hong Kong University of Science and Technology, Hong Kong }\\
\small{Email: cuiying@ust.hk, tm$\_$hqxaa@stu.ust.hk,
eeknlau@ust.hk}} }

\maketitle

\begin{abstract}
In this paper, we propose a two-timescale delay-optimal dynamic
clustering and power allocation design for downlink network MIMO
systems. The dynamic clustering control is adaptive to the global
queue state information (GQSI) only and computed at the base station
controller (BSC) over a longer time scale. On the other hand, the
power allocations of all the BSs in one cluster are adaptive to both
intra-cluster channel \textcolor{black}{state} information (CCSI)
and intra-cluster queue state information (CQSI), and computed at
the cluster manager (CM) over a shorter time scale. We show that the
two-timescale delay-optimal control can be formulated as an
infinite-horizon average cost Constrained Partially Observed Markov
Decision Process (CPOMDP). By exploiting the special problem
structure, we shall derive \textcolor{black}{an} equivalent {\em
Bellman equation} in terms of  {\em Pattern Selection Q-factor} to
solve the CPOMDP. To address the distributive requirement and the
issue of exponential memory requirement and computational
complexity, we approximate the {\em Pattern Selection Q-factor} by
the sum of {\em Per-cluster Potential} functions and propose a novel
distributive online learning algorithm to estimate the {\em
Per-cluster Potential} functions (at each CM) as well as the
Lagrange multipliers (LM) (at each BS). We show that the proposed
distributive online learning algorithm converges almost surely (with
probability 1). By exploiting the birth-death structure of the queue
dynamics, we further decompose the {\em Per-cluster Potential}
function into sum of {\em Per-cluster Per-user Potential} functions
and formulate the instantaneous power allocation as a {\em Per-stage
QSI-aware Interference Game} played among all the CMs. We also
propose a {\em QSI-aware Simultaneous Iterative Water-filling
Algorithm} (QSIWFA) and show that it can achieve the Nash
Equilibrium (NE).
\end{abstract}

\newpage

\section{Introduction}\label{sec:intro}


The network MIMO/Cooperative MIMO system is proposed as one
effective solution to address the inter-cell interference (ICI)
bottleneck in multicell systems by exploiting data cooperation and
joint processing among multiple base stations (BS). Channel state
information (CSI) and user data exchange among BSs through the
backhaul are required to support network MIMO and this overhead
depends on the number of BSs involved in the cooperation and joint
processing. In practice, it is not possible to support such
full-scale cooperation and BSs are usually grouped into disjoint
{\em clusters} with limited number of BSs in each cluster to reduce
the processing complexity as well as the backhaul loading. The BSs
within each cluster cooperatively serve the users associated with
them, which lowers the system complexity and completely eliminate
the intra-cluster interference.

The clustering methods can be classified into two categories: static
clustering approach and dynamic clustering approach. For static
clustering, the clusters are pre-determined and do not change over
time. \textcolor{black}{For example, in
\cite{Howard2007:networkmimo},\cite{RobertHeath2009:networkmimo},
the authors proposed BS coordination strategies for fixed clusters
to eliminate intra-cluster interference.}
For dynamic clustering, the cooperation clusters change in time. For
example, in \cite{Belllab2008:networkmimoclustering}, given GCSI, a
central unit jointly forms the clusters, selects the users and
calculates the beamforming coefficients and the power allocations to
maximize the weighted sum rate by a brute force exhaustive search.
In \cite{Gesbert2008:dynamicclustering}, the authors proposed a
greedy dynamic clustering algorithm to improve the sum rate under
the assumption that CSI of the neighboring BSs 
is available at each BS. 
In general, compared with static clustering, the dynamic clustering
approach usually has better performance due to larger optimizing
domain, while it also leads to larger signaling overhead to obtain
more CSI and higher computational complexity for intelligent
clustering.

\begin{figure}[h]
 \begin{center}
 \includegraphics[width=5cm]{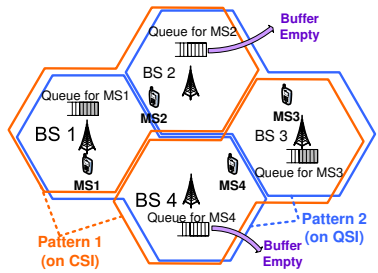}
 \caption{Motivating Example of the advantage of QSI-based Clustering
 versus traditional CSI-based Clustering in Network MIMO Systems. For
  example, CSI-based clustering will always choose Clustering Pattern
  1 (Red).  This will create an interference profile in favor of MS2
  and MS4 regardless of the queue state in MS2 and MS4. When the
  queues of MS2 and MS4 are empty, Clustering Pattern 1 will no longer
  be a good choice. On the contrary, the QSI-based clustering method
  will choose between Clustering Pattern 1 (red) and Clustering
  Pattern 2 (Blue) based on the queue states of the mobiles. As a
  result, it could dynamically creates a favorable interference
  profile to selected mobiles based on their queue
  states.}\label{fig:motivation}
  \end{center}
  \end{figure}

However, all of these works have assumed that there are infinite
backlogs of packets at the transmitter and assume the information
flow is delay insensitive. The control policy derived (e.g.
clustering and power allocation policy) is only a function of CSI
explicitly or implicitly. In practice, a lot of applications are
delay sensitive, and it is critical to optimize the delay
performance for the network MIMO systems. In particular, we are
interested to investigate delay-optimal clustering and power control
in network MIMO systems, 
which also adapts to the queue state information (QSI)
. This is motivated by an example in Fig. \ref{fig:motivation}. The
CSI-based clustering will always pick Pattern 1, creating a
cooperation and interference profile in favor of MS 2 and MS 4
regardless of the queue states of these mobiles. However, the
QSI-based clustering will dynamically pick the clustering patterns
according to the queue states of all the mobiles.

The design framework taking consideration of queueing delay and
physical layer performance is not trivial as it involves queuing
theory (to model the queuing dynamics) and information theory (to
model the physical layer dynamics). The simplest approach is to
convert the delay constraints into an equivalent average rate
constraint using tail probability (large derivation theory) and
solve the optimization problem using purely information theoretical
formulation based on the rate constraint \cite{Hui:2007}. However,
the control policy derived is a function of the CSI only, and it
failed to exploit the QSI in the adaptation
process. 
Lyapunov drift approach is also widely used in the literature to
study the queue stability region of different wireless systems and
establish throughput optimal control policy (in stability sense).
However, the average delay bound derived in terms of the Lyapunov
drift is tight only for heavy traffic loading\cite{Neelybook:2006}.
A systematic approach in dealing with delay-optimal resource control
in general delay regime is via Markov Decision Process (MDP)
technique\cite{Bertsekas:2007}. However, there are various technical
challenges involved regarding dynamic clustering and power
allocation for delay-optimal network MIMO systems.


\begin{itemize}

\item {\bf The Curse of Dimensionality:} Although MDP technique is the systematic approach to solve
the delay-optimal control problem, a first order challenge is the
curse of dimensionality\cite{Bertsekas:2007}. For example, a huge
state space (exponential in the total number of users in the
network) will be involved in the MDP and brute force value or policy
iterations cannot lead to any implementable solutions
\cite{ADP:2007}\footnote{For a multi-cell system with 7 BSs, 2 users
served by each BS, a buffer size of 10 per user and 50 CSI states
for each link between one user and one BS, the system state space
contains $(10+1)^{2\times7}\times 50^{ 7 \times 2\times7}$ states,
which is already unmanageable.}. 

\item {\bf Signaling Overhead and Computational Complexity for Dynamic Clustering:} Optimal dynamic clustering in
\cite{Belllab2008:networkmimoclustering} and greedy dynamic
clustering in \cite{Gesbert2008:dynamicclustering} (both in
throughput sense) require GCSI or CSI of all neighboring BSs, which
leads to heavy signaling overhead on backhaul and high computational
complexity for the central controller.
For delay-optimal network MIMO control, the entire system state is
characterized by the GCSI  and the global QSI (GQSI).
Therefore,  the centralized solution (which requires GCSI and GQSI)
will induce substantial signaling overhead between the BSs and the
base station controller (BSC).

\item {\bf Issues of Convergence in Stochastic Optimization Problem:}
In conventional iterative solutions for deterministic
\textcolor{black}{network utility maximization (NUM)} problems, the
updates in the iterative algorithms (such as subgradient search) are
performed within the coherence time of the CSI (the CSI remains
quasi-static during the iteration updates)\footnote{This poses a
serious limitation on the practicality of the distributive iterative
solutions because the convergence and the optimality of the
iterative solutions are not guaranteed if the CSI changes
significantly during the update.}
\cite{PalomarMungdecomposition:2006}. When we consider the
delay-optimal problem, the problem is stochastic and the control
actions are defined over the ergodic realizations of the system
states (CSI,QSI). Therefore, the convergence proof is also quite
challenging. 
\end{itemize}

In this paper, we consider a two-timescale delay-optimal dynamic
clustering and power allocation for the downlink network MIMO
consisting of $B$
cells with one BS and $K$ MSs in each cell. 
For implementation consideration, the dynamic clustering control is
adaptive to the GQSI only and computed at the BSC over a longer time
scale. On the other hand, the power allocations of all the BSs in
one cluster are adaptive to both CCSI and intra-cluster QSI (CQSI),
and computed at the CM over a shorter time scale. Due to the two
time-scale control structure, the delay optimal control is
formulated as an infinite-horizon average cost Constrained Partially
Observed Markov Decision Process (CPOMDP).
We propose an \textcolor{black}{{\em equivalent Bellman equation}}
in terms of {\em Pattern Selectio Q-factor} to solve the CPOMDP.
\textcolor{black}{We} approximate the {\em Pattern Selection
  Q-factor} by the sum of {\em Per-cluster Potential} functions and
propose a novel distributive online learning algorithm to estimate
the {\em Per-cluster Potential} functions (at each CM) as well as
the Lagrange multipliers (LM) (at each BS). This update algorithm
requires CCSI and CQSI only and therefore, facilitates distributive
implementations. Using separation of time scales, we shall establish
the almost-sure convergence proof of the proposed distributive
online learning algorithm. By exploiting the birth-death structure
of the queue dynamics, we further decompose the {\em Per-cluster
Potential} function into sum of {\em Per-cluster Per-user Potential}
functions.  Based on these distributive potential functions and
birth-death structure, the instantaneous power allocation control is
formulated as a {\em Per-stage QSI-aware Interference Game} and
determined by a {\em QSI-aware Simultaneous Iterative Water-filling
Algorithm} (QSIWFA). We show that QSIWFA can achieve the NE of the
QSI-aware interference game. Unlike conventional iterative
water-filling solutions \cite{PalomarmuMIMOunifiedview:2008}, the
{\em water-level} of our solution is adaptive to the QSI via the
potential functions.

\textcolor{black}{We first list the acronyms used in this paper in
Table \ref{table:acronyms}:}
  \begin{table}[h]
    \centering
     \textcolor{black}{
    \begin{tabular}{|l|l||l|l|}
  \hline
  BSC& base station controller & CM & cluster manager\\
  ICI& inter-cell interference & LM & Lagrange multiplier\\
  \hline
  L/C/G CSI (QSI) &
  \multicolumn{3}{l|}{local/intra-cluster/global channel state information (queue state information)}\\
  CPOMDP &
  \multicolumn{3}{l|}{constrained partially observed Markov decision
    process}\\
  QSIWFA &
  \multicolumn{3}{l|}{QSI-aware simultaneous iterative water-filling
    algorithm }\\
  \hline
\end{tabular}
}
  \label{table:acronyms}
    \caption{List of Acronyms}
    \label{tab:acronym}
  \end{table}


\section{System Models}\label{sec:system_model}
In this section, we shall elaborate the network MIMO system
topology, the physical layer model, the bursty source model and the
control policy.


\subsection{System Topology}
\begin{figure}[h]
  \begin{center}
    \subfigure[System Model]
    {\resizebox*{4cm}{5cm}{\includegraphics{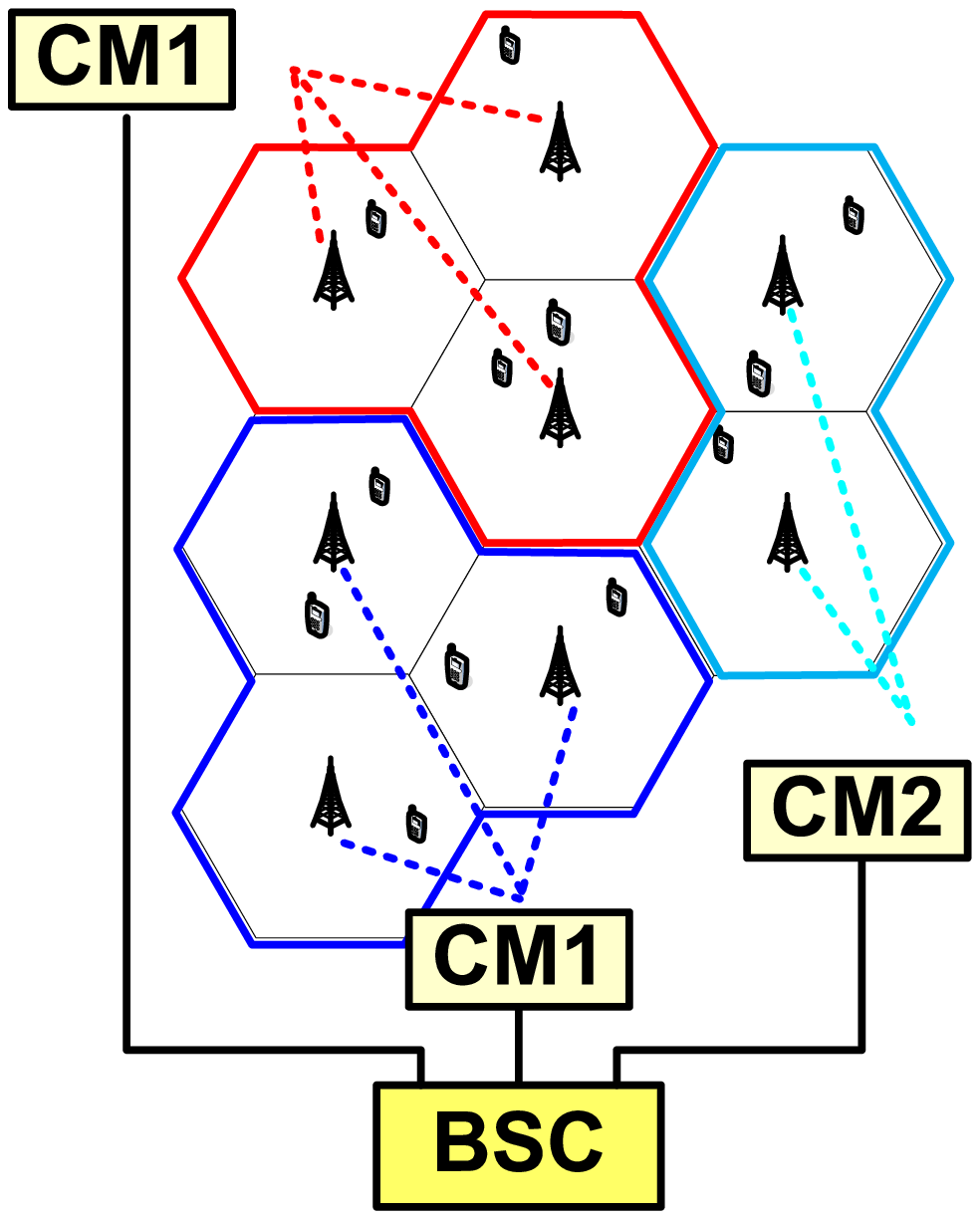}}}
    \subfigure[Control Architecture]
    {\resizebox*{12cm}{5cm}{\includegraphics{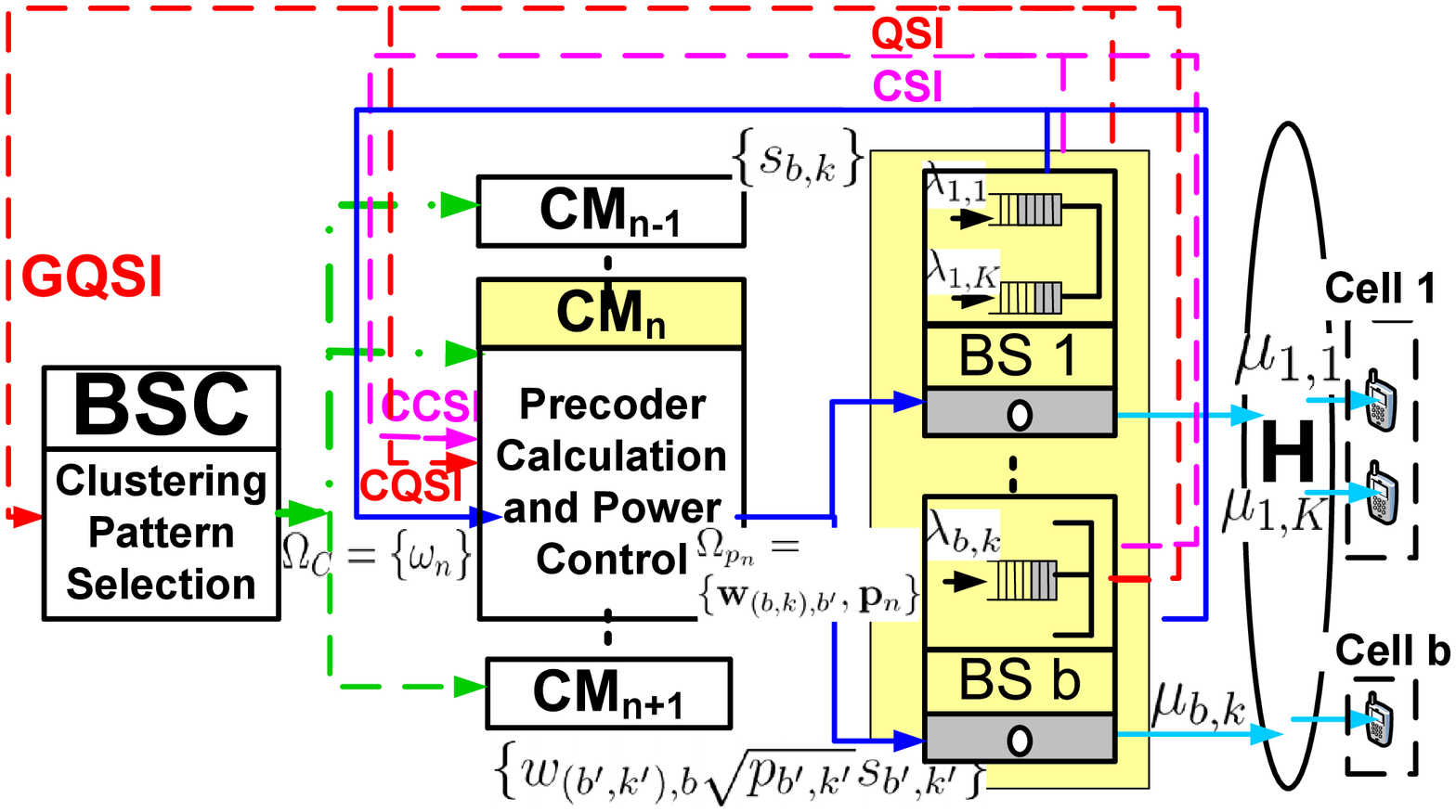}}}
    \end{center}
    \caption{
      System model and control architecture of network MIMO systems. The dotted lines and solid lines on Fig 2. (b) denote the control path and data path, respectively.}
    \label{fig:system-model}
  \end{figure}

We consider a wireless cellular network consisting of $B$ cells with
one BS and $K$ MSs in each cell as illustrated in Fig.
\ref{fig:system-model}. We assume each BS is equipped with
$\textcolor{black}{N_t \geq K}$ transmitter antennas and each MS has
$1$ receiver antenna\footnote{\textcolor{black}{When $N_t < K$,
there will be a user selection control to select at most $N_t$
active users from the $K$ users and the proposed solution framework
could be extended easily to accommodate this user selection control
as well.}}.
Denote the set of $B$ BSs as $\mathcal{B} =\{1,\cdots, B\}$ and the
set of $K$ MSs in each cell as $\mathcal{K}=\{1,\cdots, K\}$,
respectively.  We consider a clustered network MIMO system with
maximum cluster size $N_B$. Let $\omega_n \subseteq\mathcal B$
denote a feasible cluster $n$, which
is a collection of $|\omega_n|$ neighboring BSs.
We define a clustering pattern $C\in \mathcal C$ to be a partition of $\mathcal
B$ as follows
\begin{align}
C = \{\omega_{n}\subseteq \mathcal{B}: \omega_{n}\cap\omega_{n'} =
\emptyset\ \forall n\neq n', \quad \cup_{\omega_n \in C}\omega_{n} =
\mathcal{B}\} \label{eqn:cluster_pattern}
\end{align}
where $\mathcal{C}$  is the collection of all clustering patterns,
with cardinality $I_C$.

As illustrated in Fig. \ref{fig:system-model}, the overall multicell
network is specified by three-layer hierarchical architecture, i.e.
the base station controller (BSC), the cluster managers (CM) and the
BSs. There are $K$ user queues at each BS, which buffer packets for
the $K$ MSs in each cell. Both the local CSI (LCSI) and local QSI
(LQSI) are measured locally at each BS. The BSC obtains the global
QSI (GQSI) from the LQSI distributed at each BS, determines the
clustering pattern according to the GQSI, and informs the CMs of the
concerned clusters with their intra-cluster QSI (CQSI). During each
scheduling slot, the CM of each cluster determines the precoding
vectors as well as the transmit power of the BSs in the cluster.


\subsection{Physical Layer Model}
Denote MS $k$ in \textcolor{black}{cell} $b$ as a BS-MS index pair
$(b, k)$. The channel from the transmit antennas in BS $b'$ to the
MS $(b,k)$ is denoted as the $1\times N_t$ vector $\mathbf
h_{(b,k),b'}$ ($\forall b, b' \in \mathcal B, k \in \mathcal K $),
with its $i$-th element ($1 \leq i \leq N_t$)
\textcolor{black}{$h_{(b,k),b'}(i)\in\mathcal H$ a discrete random
variable distributed according to a general distribution
$P_{h_{(b,k),b'}}(h)$ with mean 0 and variance $\sigma_{(b,k),b'}$,
where $\mathcal H$ denotes the per-user discrete CSI state space
with cardinality $N_H$ and $\sigma_{(b,k),b'}$ denotes the path
gain} between BS $b'$ and MS $(b,k)$. For a given clustering pattern
$C$, let $\mathbf H_{b,n}=\{\mathbf h_{(b,k),b'}:b' \in \omega_n, k
\in \mathcal K\}$ ($\forall \omega_n \in C, b \in \omega_n$),
$\mathbf H_{n}=\cup_{b\in \omega_n}\mathbf H_{b,n}$ ($\forall
\omega_n \in
C$) 
and $\mathbf H=\cup_{\omega_n\in C}\mathbf H_n\textcolor{black}{\in
\boldsymbol{\mathcal{H}}}$ denote the LCSI at BS $b$ in cluster $n$,
the CCSI at the CM
$n$, 
and the GCSI, respectively, \textcolor{black}{where
$\boldsymbol{\mathcal{H}}$ denotes the GCSI state space}. In this
paper, the time dimension is partitioned into scheduling slots
indexed by $t$ with slot duration $\tau$.
\begin{Asump}
The GCSI $\mathbf H(t)\textcolor{black}{\in
\boldsymbol{\mathcal{H}}}$ is quasi-static in each scheduling slot
\textcolor{black}{and i.i.d. over scheduling slots. Furthermore,
$\textcolor{black}{h_{(b,k),b'}(t)\in\mathcal H}$ is independent
w.r.t. $\{(b,k),b'\}$ and $t$.} The \textcolor{black}{path gain}
$\sigma_{(b,k),b'}$ remains constant for the duration of the
communication session. 
~ \hfill\QED \label{Asump:H}
\end{Asump}

Let $s_{b,k}$ and $p_{b,k}$ ($\forall b\in \mathcal B, k \in
\mathcal K $) denote the information symbols and the received power
of MS $(b,k)$, respectively. Denote $\mathbf w_{(b,k),b'}$ ($\forall
b,b'\in\omega_n$) as  the $N_t\times 1$ precoding vector for  MS
$(b,k)$ at the BS $b'$. Therefore, the received signal of MS $(b,k)$
in cluster $n$ ($\omega_n \in C$) is given by
\begin{align}
\mathbf{y}_{b,k} =
&\underbrace{(\sum_{b'\in\omega_n}\mathbf{h}_{(b,k),b'}\mathbf
w_{(b,k),b'} )\sqrt{p_{b,k}}s_{b,k}}_{\text{desired signal}} +
\underbrace{ \sum_{\substack{b''\in\omega_n,k''\in\mathcal
K\\(b'',k'')\neq(b,k)}} (\sum_{b'\in\omega_n}
\mathbf{h}_{(b,k),b'}\mathbf w_{(b'',k''),b'})
\sqrt{p_{b'',k''}}s_{b'',k''}}_{\text{intra-cluster interference}}
\nonumber\\
 &+ \underbrace{\sum_{\substack{\omega_{n'}\in C\\ n'\neq n}}\sum_{\substack{b''\in\omega_{n'}\\ k''\in\mathcal K }} (\sum_{b'\in\omega_{n'}}
 \mathbf{h}_{(b,k),b'}\mathbf w_{(b'',k''),b'})
 \sqrt{p_{b'',k''}}s_{b'',k''}}_{\text{inter-cluster interference}}
+ \underbrace{z_{b,k}}_{\text{noise}}, \ \forall b \in \omega_n, k
\in \mathcal K, \omega_n \in C \nonumber
\end{align}
where $z_{b,k}\sim \mathcal{CN}(0,1)$ is noise. Based on CCSI at the
CM, we adopt zero-forcing (ZF) within each cluster to eliminate the
intra-cluster interference\footnote{We consider ZF precoding as an
example but the solution framework in the paper can be applied to
other SDMA processing techniques as well. \textcolor{black}{Our
zero-forcing precoder design can also be extended for multi-antenna
MS with block zero-forcing similar to that in
\cite{spencer2004zero}.}}\cite{Howard2007:networkmimo,Belllab2008:networkmimoclustering}.
The ZF precoder of cluster $n$ ($\omega_n \in C$) $\{\mathbf
w_{(b,k),b'}:b,b'\in \omega_n, k \in \mathcal K\}$ satisfies
$\sum_{b'\in\omega_n}\mathbf{h}_{(b,k),b'}\mathbf w_{(b,k),b'} =1$
($ \forall b\in\omega_n,k\in\mathcal K,\omega_n \in C$) and
$\sum_{b'\in\omega_n} \mathbf{h}_{(b,k),b'}\mathbf
w_{(b'',k''),b'}=0$ ($\forall b,b''\in\omega_n,\  k,k''\in\mathcal
K,\ (b'',k'')\neq(b,k)$). 
The  transmit power of BS $b$ is therefore given by
\begin{align}
P_{b}&= \sum_{b'\in\omega_n} \sum_{k\in\mathcal K}{\parallel
\mathbf{ w}_{(b',k),b}\parallel^2 p_{b',k}} ,\ \forall b\in\mathcal
B \label{eqn:per-BS-tx-pwr}
\end{align}


For simplicity, we assume perfect CSI at the transmitter and
receiver, and the maximum achievable data rate (bit/s/Hz) of MS
$(b,k)$ in cluster $\omega_n$ is given by the mutual information
between the channel inputs $s_{b,k}$ and channel outputs $y_{b,k}$
as:
\begin{align}
R_{b,k} = \log(1+\text{SINR}_{b,k})=\log \big(1+\frac{p_{b,k}}
{1+I_{b,k}} \big), \ \forall b \in \omega_n, k \in \mathcal K,
\omega_n \in C\label{eqn:data_rate_per_user}
\end{align}
where $\textcolor{black}{I_{b,k}}=\sum_{\substack{\omega_{n'}\in C\\
n'\neq n}}\sum_{\substack{b''\in\omega_{n'}\\ k''\in\mathcal K }}
(\sum_{b'\in\omega_{n'}} \big|\mathbf{h}_{(b,k),b'}\mathbf
w_{(b'',k''),b'}\big|^2) p_{b'',k''}$ ($\forall b \in \omega_n, k
\in \mathcal K, \omega_n \in C$) is the inter-cell interference
power.

\subsection{Bursty  Source Model}
Let $\mathbf{A}(t)=\{A_{b,k}(t): b\in\mathcal{B}, k\in
\mathcal{K}\}$  be the  random new arrivals (number of bits) for the
$BK$ users in the multicell network at the end of the $t$-th
scheduling slot.
\begin{Asump}
The arrival process  $A_{b,k}(t)$ is distributed according to
general distributions $P_{A_{b,k}}(A)$ and is i.i.d. over scheduling
slots and independent w.r.t. $\{(b,k)\}$. ~ \hfill\QED
\label{Asump:general_A-N}
\end{Asump}


Let $\mathbf{Q}(t)\textcolor{black}{\in\boldsymbol{\mathcal Q}}$ be
the $B\times K$ GQSI matrix of the multicell network, where
$\textcolor{black}{Q_{b,k}(t)\in \mathcal Q}$ is the $(b,k)$-element
of $\mathbf{Q}(t)$, which denotes the number of bits in the queue
for MS $(b,k)$ at the beginning of the $t$-th slot.
\textcolor{black}{The per-user QSI state space and the GQSI
  state space are given by $\mathcal Q=\{0,1,\cdots, N_Q\}$, and
  $\boldsymbol{\mathcal Q}= \mathcal Q^{BK}$}, separately. $N_Q$ denotes the buffer size (maximum number of
bits) of the queues for the $BK$ MSs.  Thus, the cardinality of the
GQSI state space is $I_Q=(N_Q+1)^{BK}$, which grows exponentially
with $BK$. Let $\mathbf{R}(t)$ be the $B\times K$ scheduled data
rates matrix of the $BK$ MSs, where the $(b,k)$-element $R_{b,k}(t)$
can be calculated using \eqref{eqn:data_rate_per_user}. We assume
the controller is causal so that new arrivals $\mathbf{A}(t)$
\textcolor{black}{are} observed only after the controller's actions
at the $t$-th slot. Hence, the queue dynamics is given by the
following equation:
\begin{align}
Q_{b,k}(t + 1) = \min \Big \{\big[Q_{b,k}(t)-R_{b,k}(t)\tau\big]^+ +
A_{b,k}(t),\ N_Q \Big\} \label{eqn:general-queue-evolution-bit}
\end{align}
where $x^+\triangleq\max\{x,0\}$ and $\tau$ is the  duration of a
scheduling slot. For notation convenience, we denote
$\boldsymbol{\chi}(t)=\big(\mathbf{H}(t),\mathbf{Q}(t)\big)$ as the
{\em global system state} at the $t$-th slot.

\subsection{Clustering Pattern Selection and Power Control Policy}

At the beginning of the $t$-th slot, given the observed GQSI
realization $\mathbf{Q}(t)$,  the BSC determines the clustering
pattern $C$ defined in \eqref{eqn:cluster_pattern},  the CMs of the
active clusters $n$ ($\forall \omega_n\in C$) do power allocation
based on GCSI and GQSI according to a {\em pattern selection and
power allocation policy} defined below.

\begin{Def}[Stationary Pattern Selection and Power Allocation Policy]\label{Defn:feasible_policy}
A stationary pattern selection and power allocation policy
$\Omega=(\Omega_c,\Omega_p)$ is a mapping from the system state
$\boldsymbol{\chi}\textcolor{black}{\in\boldsymbol{\mathcal{X}}}$ to
the pattern selection and power allocation actions, where
$\Omega_c(\mathbf{Q})= C\in \mathcal C$ and
$\Omega_p(\boldsymbol{\chi})=\{p_{b,k}: b \in \mathcal{B}, k \in
\mathcal{K}\}$. A policy $\Omega$ is called {\em feasible} if the
associated actions satisfy the  per-BS average transmit power
constraint given by
\begin{align}
\mathbb{E}^{\Omega}[P_{b}]\leq \overline{P}_b,\ \forall b \in \mathcal{B}
\label{eqn:tx-pwr-constraint}
\end{align}
where $P_{b}$ is given by \eqref{eqn:per-BS-tx-pwr} and $\overline{P}_b$
is the average total power of BS $b$. ~ \hfill\QED
\end{Def}

\begin{Rem}[Two Time-Scale Control Policy]
  The pattern selection policy is defined as a function of GQSI
  only, i.e. $\Omega_c(\mathbf{Q})$, for the following reasons. The
  QSI is changing on a slower time scale while the CSI is changing
  on a faster (slot-by-slot) time scale. The dynamic clustering is
  enforced at the BSC and hence, a longer time scale will be
  desirable from the implementation perspective, considering
  computational complexity at the BSC and signaling overhead for
  collecting GCSI from all the BSs. \textcolor{black}{On the other hand, the low
  complexity and decentralized power allocation policy (obtained
  later in Sec. IV) is a function of CQSI and CCSI only and executed
  at the CM level distributively\footnote{\textcolor{black}{According to Definition
    1, the power control policy $\Omega_p$ is defined as a function
    of the GQSI and GCSI. Yet, in Sec.IV, we shall derive a
    decentralized power allocation policy, which is adaptive to CCSI
    and CQSI only.}}, and hence it can operate at slot-time scale
  with acceptable signaling overhead and complexity.} ~ \hfill\QED
\end{Rem}

\section{Problem Formulation}\label{sec:CMDP_formulation}
In this section, we shall first elaborate the dynamics of the system
state under a control policy $\Omega$. Based on that, we shall then
formally formulate the delay-optimal control problem for  network
MIMO systems.

\subsection{Dynamics of System State}

A stationary control policy $\Omega$ induces a joint distribution
for the random process $\{\boldsymbol{\chi}(t)\}$.  Under Assumption
\ref{Asump:H} and \ref{Asump:general_A-N}, the arrival and departure
are memoryless. Therefore, the induced random process
$\{\boldsymbol{\chi}(t)\}$ for a given control policy $\Omega$ is
Markovian with the following transition
probability:
\begin{align}
&\Pr[\boldsymbol{\chi}(t+1)|\boldsymbol{\chi}(t),\Omega(\boldsymbol{\chi}(t))]=
\Pr[\mathbf{H}(t+1)|\boldsymbol{\chi}(t),\Omega(\boldsymbol{\chi}(t))]
\Pr[\mathbf{Q}(t+1)|\boldsymbol{\chi}(t),\Omega(\boldsymbol{\chi}(t))]\nonumber\\
=&\Pr[\mathbf{H}(t+1)]
\Pr[\mathbf{Q}(t+1)|\boldsymbol{\chi}(t),\Omega(\boldsymbol{\chi}(t))]\label{eqn:transition-prob1}
\end{align}
Note that the $BK$ queues are coupled together via the control
policy $\Omega$.

\subsection{Delay Optimal Problem Formulation}
Given a unichain policy $\Omega$, the induced Markov chain
$\{\boldsymbol{\chi}(t)\}$ is ergodic\footnote{
\textcolor{black}{The unichain policy is defined as a policy under
which the resulting Markov chain is
ergodic\cite{tsitsiklis07unichain}. Similar to other literature in
MDP \cite{Bertsekas:2007},\cite{Cao:2007}, we restrict out
consideration to unchain policy in this paper. Such assumption
usually does not contribute any loss of performance. For example, in
Section \ref{sec:application_poisson_exponential}, any
non-degenerate control policy satisfies $\mathbb E[p_{b,k}(\mathbf
H,\mathbf
    Q)|\mathbf Q]>0, \forall Q_{b,k}>0$,
    i.e. $\overline\mu_{b,k}(Q_{b,k})>0, \forall Q_{b,k}>0$. Hence, the induced Markov
    chain $\{\mathbf Q(t)\}$ is an ergodic birth death process.
  }
} and there exists a unique
steady state distribution $\pi_{\chi}$ where
$\pi_{\chi}(\boldsymbol{\chi})=\lim_{t\rightarrow\infty}\Pr[\boldsymbol{\chi}(t)
= \boldsymbol{\chi}]$. The average cost of MS $(b,k)$ under a
unichain policy $\Omega$ is given by:
\begin{equation}
\overline{D}_{b,k}(\Omega) = \lim_{T\rightarrow\infty}
\frac{1}{T}\sum_{t=1}^T\mathbb{E}^{\Omega}\big[f\big(Q_{b,k}(t)\big)\big]
=\mathbb{E}_{\pi_{\chi}}\left[f(Q_{b,k})\right],\ \forall b \in
\mathcal{B}, k \in \mathcal{K} \label{eqn:delay1}
\end{equation}
where $f(Q_{b,k})$ is a monotonic increasing utility function of
$Q_{b,k}$ and the $\mathbb{E}_{\pi_{\chi}}$ denotes expectation
w.r.t. the underlying measure $\pi_{\chi}$. For example, when
$f(Q_{b,k})=\frac{Q_{b,k}}{\lambda_{b,k}}$,
$\overline{D}_{b,k}(\Omega)
=\frac{\mathbb{E}_{\pi_{\chi}}\left[Q_{b,k}\right]}{\lambda_{b,k}} $
is the {\em average delay} of MS $(b,k)$ (by Little's Law). Another
interesting example is the {\em queue outage probability}
$\overline{D}_{b,k}(\Omega)=\Pr[Q_{b,k}\geq Q_{b,k}^o]$, in which
$f(Q_{b,k})=\mathbf{1}[Q_{b,k}\geq Q_{b,k}^o]$, where
$\textcolor{black}{Q_{b,k}^o \in \mathcal Q}$ is the reference
outage queue state.
Similarly, the average transmit power constraint in
(\ref{eqn:tx-pwr-constraint})  can be written as
\begin{align}
\overline{P_b}(\Omega)=\lim_{T\rightarrow\infty}\frac{1}{T}\sum_{t=1}^T\mathbb{E}^{\Omega}[P_{b}(t)]=\mathbb{E}_{\pi_{\chi}}[P_b]\leq
\overline{P}_b,\ \forall b \in \mathcal{B} \label{eqn:tx-pwr2}
\end{align}
where $P_{b}$ is given by \eqref{eqn:per-BS-tx-pwr}.

In this paper, we seek to find an optimal stationary
\textcolor{black}{unichain} control policy to minimize the average
cost in \eqref{eqn:delay1}. Specifically, we have
\begin{Prob}[Delay-Optimal Control Problem for Network MIMO]
For some positive constants\footnote{The positive weighting factors
$\boldsymbol{\beta}$ in \eqref{cons-MDP} indicate the relative
importance of buffer delay among the $BK$ data streams and for each
given $\boldsymbol{\beta}$, the solution to \eqref{cons-MDP}
corresponds to a point on the Pareto optimal delay tradeoff boundary
of a {\em multi-objective} optimization problem.}
$\boldsymbol{\beta}=\{\beta_{b,k}>0:n\in \mathcal{B}, k\in
\mathcal{K}\}$, the delay-optimal problem is formulated as
\begin{eqnarray}
& &
\min_{\Omega}J_{\beta}(\Omega)=\sum_{b,k}\beta_{b,k}\overline{D}_{b,k}(\Omega)=
\lim_{T\rightarrow\infty} \frac{1}{T}\sum_{t=1}^T
\mathbb{E}^{\Omega}\Big[
d\big(\boldsymbol{\chi}(t),\Omega(\boldsymbol{\chi}(t))\big)\Big]
\label{cons-MDP}\\
& & \text{subject to} \quad \text{the power constraints in}
\,(\ref{eqn:tx-pwr2})\nonumber
\end{eqnarray}
where
$d\big(\boldsymbol{\chi}(t),\Omega(\boldsymbol{\chi}(t))\big)=\sum_{b,k}
\beta_{b,k} f\big(Q_{b,k}(t)\big)$. ~ \hfill\QED \label{Prob2}
\end{Prob}

\subsection{Constrained POMDP}

Next, we shall illustrate that Problem \ref{Prob2} is an infinite
horizon average cost constrained POMDP. 
In Problem \ref{Prob2}, the pattern selection policy is defined on
the partial system state $\mathbf Q$, while the power allocation
policy is defined on the complete system state $\boldsymbol \chi
=(\mathbf H, \mathbf
Q)\textcolor{black}{\in\boldsymbol{\mathcal{X}}}$\textcolor{black}{,
  where $\boldsymbol{\mathcal{X}}=\boldsymbol{\mathcal{H}}\times\boldsymbol{\mathcal{Q}}$}. Therefore,  Problem \ref{Prob2} is a
constrained POMDP (CPOMDP) with the following specification:
\begin{itemize}
\item \textbf{State Space}: The state space is given by \textcolor{black}{ $\{ (\mathbf{H},\mathbf{Q}):\forall \mathbf{H}\in\boldsymbol{\mathcal{H}}, \mathbf{Q}\in\boldsymbol{\mathcal{Q}}\}$, where $(\mathbf{H},\mathbf{Q})$} is a realization of
the global system state.
\item \textbf{Action Space}: The action space  is given by \textcolor{black}{$\{\Omega(\mathbf{H},\mathbf{Q}):\forall \mathbf{H}\in\boldsymbol{\mathcal{H}}, \mathbf{Q}\in\boldsymbol{\mathcal{Q}}\}$}, where $\Omega=(\Omega_c,\Omega_p)$ is a unichain
feasible policy as defined in Definition \ref{Defn:feasible_policy}.
\item \textbf{Transition Kernel}: The transition kernel  $\Pr[\boldsymbol{\chi}'|\boldsymbol{\chi},\Omega(\boldsymbol{\chi})]$ is given by \eqref{eqn:transition-prob1}.
\item \textbf{Per-stage Cost Function}: The per-stage cost function is given by $d(\boldsymbol{\chi},\Omega(\boldsymbol{\chi}))=\sum_{b,k} \beta_{b,k} f(Q_{b,k})$.
\item \textbf{Observation}: The observation for the pattern selection policy is GQSI, i.e.,
$o_c = \mathbf Q$, while the observation for the power allocation
policy is the complete system state, i.e. $o_p=\boldsymbol \chi$.
\item \textbf{Observation Function}: The observation function for the pattern selection policy is\\
$\textcolor{black}{O_c\big(o_c, \boldsymbol \chi, (\Omega_c(\mathbf
Q'), \Omega_p(\boldsymbol \chi'))\big) = 1}$, if $o_c = \mathbf Q$,
otherwise 0. Similarly, the observation function for the power
allocation policy is $\textcolor{black}{O_p\big(o_p, \boldsymbol
\chi, (\Omega_c(\mathbf Q'), \Omega_p(\boldsymbol \chi'))\big) =
1}$, if $o_p =\boldsymbol \chi$, otherwise 0.
\end{itemize}

For any  Lagrangian multiplier (LM) vector
$\boldsymbol{\gamma}=(\gamma_1,\cdots,\gamma_B)$ ($\forall \gamma_b
\geq 0$), define the Lagrangian as
\[
L_{\beta}(\Omega,\boldsymbol{\gamma})=\lim_{T\rightarrow\infty}
\frac{1}{T}\sum_{t=1}^T \mathbb{E}^{\Omega}\Big[
g\big(\boldsymbol{\gamma},\boldsymbol{\chi}(t),\Omega(\boldsymbol{\chi}(t))\big)\Big]
\]
where
$g(\boldsymbol{\gamma},\boldsymbol{\chi},\Omega(\boldsymbol{\chi}))
= \sum_{b \in \mathcal B} \Big(\sum_{k\in \mathcal K}\beta_{b,k}
f(Q_{b,k})+ \gamma_b(P_{b}-\overline{P}_b)\Big)$. Therefore, the
corresponding unconstrained POMDP for a particular LM vector
$\boldsymbol{\gamma}$ \textcolor{black}{(i.e. the Lagrange dual
function)} is given by
\begin{eqnarray}
G(\boldsymbol{\gamma})=\min_{\Omega}L_{\beta}(\Omega,\boldsymbol{\gamma})
\label{uncons-MDP}
\end{eqnarray}
\textcolor{black}{The dual problem of the primal problem in Problem
\ref{Prob2} is given by
$\max_{\boldsymbol{\gamma}\succeq0}G(\boldsymbol{\gamma})$.  It is
shown in \cite{Borkarbook:2008} that there exists a Lagrange
multiplier $\boldsymbol{\gamma}\ge 0$ such that $\Omega^*$ minimizes
$L_{\beta}(\Omega,\boldsymbol{\gamma})$ and the saddle point
condition $L_{\beta}(\Omega^*, \boldsymbol{\gamma})\leq
L_{\beta}(\Omega^*, \boldsymbol{\gamma}^*) \leq L_{\beta}(\Omega,
\boldsymbol{\gamma}^*)$ holds, i.e.
$(\Omega^*,\boldsymbol{\gamma}^*)$ is a saddle point of the
Lagrangian $L_{\beta}(\Omega, \boldsymbol{\gamma})$. Using standard
optimization theory\cite{boydconvex:2004}, $\Omega^*$ is the primal
optimal (i.e. the optimal solution of the original Problem 1),
$\gamma^*$ is the dual optimal (i.e. the optimal solution of the
dual problem), and the duality gap (i.e. the gap between the primal
objective at $\Omega^*$ and the dual objective at $\gamma^*$) is
zero. Therefore, by solving the dual problem, we can obtain the
primal optimal $\Omega^*$. }

\subsection{ \textcolor{black}{Equivalent Bellman Equation}}

While POMDP is a very difficult problem in general, we shall exploit
some special structures in our problem to substantially simplify the
problem. We first define  conditional power allocation action sets
below:

\begin{Def}[Conditional Power Allocation Action Sets]
Given a power allocation policy $\Omega_p$, we define a conditional
power allocation set $\Omega_p(\mathbf{Q})=\{\mathbf{p} =
\Omega_p(\boldsymbol{\chi}):
\boldsymbol{\chi}=(\mathbf{Q},\mathbf{H}), \forall \mathbf{H}\}$ as
the collection of actions for all possible CSI $\mathbf{H}$
conditioned on a given QSI $\mathbf{Q}$. The complete control policy
$\Omega_p$ is therefore equal to the union of all the conditional
power allocation action sets. i.e. $\Omega_p =
\bigcup_{\mathbf{Q}}\Omega_p(\mathbf{Q})$.
\label{defn:conditional-action1} ~ \hfill\QED
\end{Def}

Based on Definition \ref{defn:conditional-action1}, we can transform
the POMDP problem into a regular infinite-horizon average cost MDP.
Furthermore, for a given $\boldsymbol \gamma$, the optimal control
policy $\Omega^*$ can be obtained by solving an
\textcolor{black}{equivalent Bellman equation} which is summarized
in the lemma below.

\begin{Lem}[\textcolor{black}{Equivalent Bellman Equation} and Pattern Selection
  Q-factor]\label{Lem:reduced-MDP}
  \textcolor{black}{ For a given LM $\boldsymbol \gamma$, the optimal
    control policy $\Omega^*=(\Omega_c^*, \Omega_p^*)$ for the
    unconstrained optimization problem in Problem \ref{Prob2} can be
    obtained by solving the following \textcolor{black}{{\em
        equivalent Bellman equation}}: ($\forall \mathbf
    Q\in\boldsymbol{\mathcal Q}, \forall C \in \mathcal{C}$)
\begin{align}
\mathbb Q(\mathbf Q,C) =\min_{\Omega_p(\mathbf{Q})}\Big\{
\overline{g}\big(\boldsymbol{\gamma},\mathbf{Q},C,\Omega_p(\mathbf{Q})\big)
+ \sum_{\mathbf{Q}'}& \Pr[\mathbf{Q}'|
\mathbf{Q},C,\Omega_p(\mathbf{Q})]
 \min_{C'}\mathbb Q(\mathbf Q',C') \Big\} -\theta\label{eqn:Bellman3}
\end{align}
where $\{\mathbb Q(\mathbf Q,C)\}$ is the {\em Pattern Selection
  Q-factor},
$\overline{g}\big(\boldsymbol{\gamma},\mathbf{Q},C,\Omega_p(\mathbf{Q})\big)=\mathbb
E
[g(\boldsymbol{\gamma},(\mathbf{H},\mathbf{Q}),C,\Omega_p(\mathbf{H},\mathbf{Q}))$
$|\mathbf{Q}]$ is the conditional per-stage cost, $\Pr[\mathbf{Q}'|
\mathbf{Q},C, \Omega_p(\mathbf{Q})]=\mathbb E \big[\Pr[\mathbf{Q}'|
(\mathbf{H},\mathbf{Q}),C, \Omega_p(\mathbf{H},\mathbf{Q})]\big
|\mathbf{Q}\big]$ is the conditional expectation of transition
kernel. If there is a $(\theta, \{\mathbb Q(\mathbf Q,C)\})$ that
satisfies the fixed-point equations in (\ref{eqn:Bellman3}), then
$\theta = L_{\beta}^*(\boldsymbol{\gamma}) =
\min_{\Omega}L_{\beta}(\Omega,\boldsymbol{\gamma})$ is the optimal
average cost in Problem \ref{Prob2}. Furthermore, the  optimal
control policy is given by $\Omega^*(\boldsymbol\gamma)=(\Omega_c^*,
\Omega_p^*)$ with $\Omega_p^*(\mathbf{Q})$ attaining the minimum of
the R.H.S. of \eqref{eqn:Bellman3} ($\forall \mathbf
Q\in\boldsymbol{\mathcal Q}, \forall C \in \mathcal{C}$) and
$\Omega_c^*(\mathbf Q)=\arg \min_{C}\mathbb Q(\mathbf Q,C)$. ~
\hfill\QED }
\end{Lem}
\begin{proof}
Please refer to the Appendix A.
\end{proof}

\begin{Rem}
The  \textcolor{black}{equivalent Bellman equation} in
\eqref{eqn:Bellman3} is defined on the GQSI $\mathbf{Q}$ with
cardinality $I_Q$ only. Nevertheless, the optimal power allocation
policy $\Omega_p^* = \bigcup_{\mathbf{Q}}\Omega_p^*(\mathbf{Q})$
obtained by solving \eqref{eqn:Bellman3} is still adaptive to  GCSI
$\mathbf{H}$ and GQSI $\mathbf{Q}$, where $\Omega_p^*(\mathbf{Q})$
are the conditional power allocation action sets given by Definition
\ref{defn:conditional-action1}. We shall illustrate this with a
simple example below. In other words, the derived policies of the
\textcolor{black}{equivalent  Bellman equation} in
\eqref{eqn:Bellman3} solve the CPOMDP in Problem \ref{Prob2}. ~
\hfill\QED
\end{Rem}

\textcolor{black}{
\begin{Exam}
  Suppose there are two MSs with the CSI state space ${\mathcal
    H}=\{ H_1, H_2\}$ as a simple example. As a result, the global
  CSI state space is $\boldsymbol{\mathcal H}=\{H_1,H_2\}^2$ with
  cardinality $I_H=4$.  Given GQSI $\mathbf Q$, the optimal
  conditional power allocation action set $\Omega_p^*(\mathbf
  Q)=\{\mathbf p^*(\mathbf H,\mathbf Q): \mathbf H \in \boldsymbol{\mathcal H}\}$ (by Definition
  \ref{defn:conditional-action1}) for any given pattern $C$ is obtained by solving the R.H.S. of
  (\ref{eqn:Bellman3}).
  \begin{align}
     &\Omega_p^*(\mathbf Q) 
    =  \arg\min_{ (\mathbf p(\mathbf H^{(i)},\mathbf
      Q))\big|_{i=1}^{4} } \Big\{ \sum_{i=1}^4 \Pr[\mathbf H^{(i)}]
    \big(
    g(\boldsymbol{\gamma},(\mathbf{H}^{(i)},\mathbf{Q}),C,\Omega_p(\mathbf{H}^{(i)},\mathbf{Q}))\nonumber\\
    &\quad \quad \quad \quad \quad \quad \quad \quad \quad \quad\quad \quad
    + \sum_{\mathbf Q'}\Pr[\mathbf Q'|\mathbf H^{(i)},\mathbf
    Q,C,\Omega_p(\mathbf H^{(i)},\mathbf Q)]\min_{C'}\mathbb
    Q(\mathbf Q',C')\big) \Big\} \nonumber
  \end{align}
  Observe that the R.H.S. of the above equation is a decoupled
  objective function w.r.t. the variables $\{\mathbf p(\mathbf H^{(i)},\mathbf Q)\}$ and
  hence
  \begin{align}
    & \mathbf p^*(\mathbf H^{(i)},\mathbf Q)) \nonumber\\
    =& \arg\min_{ \mathbf
      p(\mathbf H^{(i)},\mathbf Q)}\Big\{
    g(\boldsymbol{\gamma},(\mathbf{H}^{(i)},\mathbf{Q}),C,\Omega_p(\mathbf{H}^{(i)},\mathbf{Q}))+
\sum_{\mathbf Q'}\Pr[\mathbf Q'|\mathbf H^{(i)},\mathbf
    Q,C,\Omega_p(\mathbf H^{(i)},\mathbf Q)]\min_{C'}\mathbb
    Q(\mathbf Q',C')
    \Big\} \nonumber
  \end{align}
Hence, using Lemma \ref{Lem:reduced-MDP}, the optimal power control
policy is given by
  $\Omega_p^* =\bigcup_{\mathbf Q}\Omega_p^*(\mathbf Q)$, which are
  functions of both the GQSI and the GCSI. The optimal
  clustering pattern selection is given by $\Omega_c^*(\mathbf
  Q)=\arg \min_{C}\mathbb Q(\mathbf Q,C)$, which is a function of
  the GQSI only.  ~\hfill\QED
  \end{Exam}
}



\section{General Low Complexity Decentralized Solution}\label{sec:general_decentral_algo}
The key steps in obtaining the optimal control policies from the
R.H.S. of the Bellman equation in \eqref{eqn:Bellman3} rely on the
knowledge of the pattern selection Q-factor $\{\mathbb Q(\mathbf
Q,C)\}$ (which involves solving a system of $I_C I_Q$ non-linear
Bellman equations in \eqref{eqn:Bellman3} for given LMs with
$I_CI_Q+1$ unknowns \textcolor{black}{($\theta,\{\mathbb Q(\mathbf
Q,C)\})$)} and the $B$ LMs $\{\gamma_b:b\in\mathcal B\}$, which
leads to enormous complexity. Brute-force solution has exponential
complexity and requires centralized implementation and knowledge of
GCSI and GQSI (which leads to huge signaling overheads). In this
section, we shall approximate the {\em pattern selection Q-factor}
by the sum of {\em Per-cluster Potential} functions. Based on the
approximation structure, we propose a novel distributive online
learning algorithm to estimate the {\em Per-cluster Potential}
functions (performed at each CM) as well as the LMs
$\{\gamma_b:b\in\mathcal B\}$ (performed at each BS).



\subsection{Linear Approximation of the Pattern Selection Q-Factor}

\textcolor{black}{Let $\boldsymbol{\mathcal Q_n}$ denote the CQSI
  state space of cluster $n$ with cardinality
  $I_{Q_{n}}=(N_Q+1)^{|\omega_n|K}$ }.  To reduce the cardinality of
the state space and to decentralize the resource allocation, we
approximate $\mathbb Q(\mathbf Q,C)$ by the sum of {\em per-cluster
potential} $\overline V_{n}(\mathbf Q_{n})$ ($\forall \omega_{n}\in
C$), i.e.
\begin{align}
\mathbb Q(\mathbf Q,C)\thickapprox \sum_{\omega_{n}\in C} \overline
V_{n}(\mathbf Q_{n}) \label{eqn:approximate-Q}
\end{align}
where \textcolor{black}{$\mathbf Q_{n}\in\boldsymbol{\mathcal Q_n}$}
is the CQSI state of cluster $n$ ($\omega_{n}\in C$) and
$\{\overline V_{n}(\mathbf Q_{n}):\forall \mathbf Q_{n}\}$
\textcolor{black}{are}  {\em per-cluster potential} functions which
satisfy the following {\em per-cluster potential fixed point
equation}: \textcolor{black}{$
  (\forall \mathcal Q_{n}\in\boldsymbol{\mathcal Q_n})$
\begin{align}
\theta_{n}+\overline V_{n}(\mathbf Q_{n}) =\min_{\Omega_{\mathbf
p_n}(\mathbf Q_{n})}\Big\{ \overline g_{n}(\boldsymbol{\gamma}_n,
\mathbf Q_{n},\Omega_{\mathbf p_n}( \mathbf Q_{n}))+
\sum_{\mathbf Q_{n}'}\Pr[\mathbf Q_{n}'| \mathbf
Q_{n},\Omega_{\mathbf p_n}( \mathbf Q_{n})] \overline V_{n}(\mathbf
Q_{n}') \Big \} \label{eqn:bellman-per-cluster-potential}
\end{align}
}
where
\begin{align}
 \overline g_{n}(\boldsymbol{\gamma}_n, \mathbf
Q_{n},\Omega_{\mathbf p_n}(\mathbf Q_{n}))=\mathbb
E\underbrace{\Big[ \sum_{b\in\omega_n}\Big(
\sum_{k}\beta_{b,k}f(Q_{b,k}) +\gamma_{b}\big(P_b(\mathbf
H_{n},\mathbf Q_{n})-\overline P_b\big)
\Big)}_{g_{n}(\boldsymbol{\gamma}_n, \mathbf H_{n},\mathbf
Q_{n},\Omega_{\mathbf p_n}(\mathbf H_{n},\mathbf Q_{n}))} \Big |
\mathbf Q_{n} \Big], \label{eqn:per-cluster-per-stage-cost}
\end{align}
\begin{align}
&\Pr[\mathbf Q_{n}'|\mathbf Q_{n},\Omega_{\mathbf p_n}(\mathbf
Q_{n})]= \mathbb E\big[\Pr[\mathbf Q_{n}'|\mathbf H_{n}, \mathbf
Q_{n},\Omega_{\mathbf p_n}(\mathbf H_{n},\mathbf Q_{n})]\big |
\mathbf Q_{n} \big],\label{eqn:general-per-cluster-tran-prob}
\end{align}
$\boldsymbol{\gamma}_n=\{\gamma_b: b \in \omega_n\}$, $P_b(\mathbf
H_{n},\mathbf Q_{n})=\sum_{b'\in\omega_n} \sum_{k}\|\mathbf {
w}_{(b',k),b}\|^2 p_{b',k}(\mathbf H_{n},\mathbf Q_{n})$ given by
\eqref{eqn:per-BS-tx-pwr},  $\mathbf p_n=\{ p_{b,k}: b \in \omega_n,
k\in \mathcal K\}$.



\textcolor{black}{In the literature, there are mainly three types of
compact representations, which can be used to approximate the
potential functions
\cite{tsitsiklis1996feature},\cite{bertsekas2002neuro}: Artificial
neural networks, Feature Extraction, and Parametric Form.  The first
two approaches still need (GCSI,GQSI), have exponential complexity
with respect to $B$ and $K$, and do not facilitate distributed
implementations. Therefore, we adopt Parametric Form with linear
approximation. Due to the cluster-based structure and the
relationship between the GQSI and the CQSI, we can extract
meaningful features and use the summation form for approximation,
which naturally lead to distributed implementation.} Using the above
linear approximation of the pattern selection Q-factor by the sum of
{\em per-cluster potential} functions in \eqref{eqn:approximate-Q},
the BSC determines the optimal clustering pattern based on the
current observed GQSI $\mathbf Q$ according to
\begin{align}
\Omega_c^*(\mathbf Q)=\arg \min_{C}\sum_{\omega_{n}\in C} \overline
V_{n}(\mathbf Q_{n}) \label{eqn:per-cluster-approximate-Q-pattern}
\end{align}
Based on the CQSI and CCSI observation $(\mathbf H_n,
\mathbf{Q}_n)$, each CM $n$ ($\omega_n \in \Omega_c^*(\mathbf Q)$)
determines $\Omega_{\mathbf p_n}^*(\mathbf{Q}_n)=\{\Omega_{\mathbf
p_n}^*(\mathbf H_n, \mathbf{Q}_n):(\mathbf H_n, \mathbf{Q}_n)\forall
\mathbf H_n\}$, which attains the minimum of the R.H.S. of
\eqref{eqn:bellman-per-cluster-potential}
\textcolor{black}{($\forall\mathbf Q\in\boldsymbol{\mathcal Q},
\forall C \in \mathcal{C}$)}. Hence, the overall power allocation
control policy is given by $\Omega_{
p}^*(\boldsymbol{\chi})=\{\Omega_{\mathbf p_n}^*(\mathbf H_n,
\mathbf{Q}_n): \omega_n \in \Omega_c^*(\mathbf Q)\}$.

\subsection{Online Primal-Dual Distributive Learning Algorithm via Stochastic Approximation}

Since the derived policy $\Omega^*=(\Omega_c^*,\Omega_{ p}^*)$ is
function of {\em per-cluster potential} functions $\{\overline
V_{n}(\mathbf Q_{n})\}$ ($\forall \omega_{n}$), we need to obtain
$\{\overline V_{n}(\mathbf Q_{n})\}$ by solving
\eqref{eqn:bellman-per-cluster-potential} and determine the LMs such
that the per-BS average power constraints in
\eqref{eqn:tx-pwr-constraint} are satisfied, which are not trivial.
In this section, we shall apply stochastic
learning\cite{Cao:2007,Abounadilearningalg:2001} to estimate the
{\em per-cluster potential} functions $\{\overline V_{n}(\mathbf
Q_{n})\}$ ($\forall \omega_{n}$) distributively at each CM based on
realtime observations of the CCSI and CQSI and LMs at each BS based
on the realtime power allocations actions. The convergence proof of
the online learning algorithm will be established in the next
section.

\begin{figure}[h]
  \begin{center}
  \includegraphics[height=7cm, width=14cm]{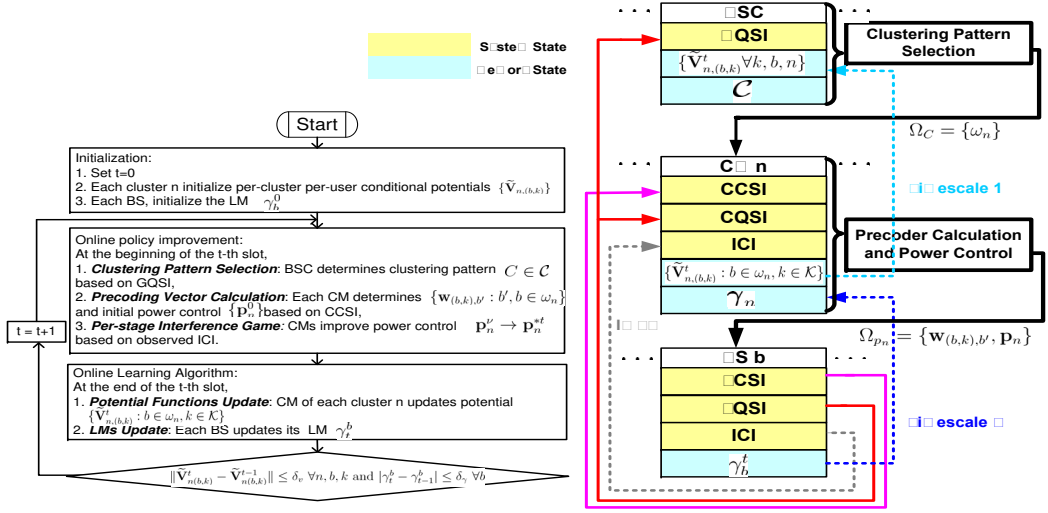}
  \caption{Algorithm Flow of the Proposed Online Distributive
  Primal-Dual Learning Algorithm with Per-stage QSI-aware Interference
  Game and Simultaneous Updates on Potential Functions and
  LMs.}\label{fig:system-flow}
  \end{center}
  \end{figure}

Fig. \ref{fig:system-flow} illustrates the top level structure of
the online learning algorithms. The {\em Online Primal-Dual
Distributive Learning Algorithm via Stochastic Approximation}, which
requires knowledge on CCSI and CQSI only, can be described as
follows:

\begin{Alg}(\emph{Online Primal-Dual Distributive Learning Algorithm via
Stochastic Approximation})

\begin{itemize}
\item {\bf Step 1 [Initialization]}: Set $t=0$. Each cluster $n$
initialize the {\em  per-cluster potential} functions
$\{V_n^0(\mathbf Q_n)\}$ ($\omega_n$). Each BS $b$ initialize the LM
$\gamma_b^0$ ($\forall b \in \mathcal B$).
\item {\bf Step 2 [Clustering Pattern Selection]}: At the beginning of the $t$-th slot, the BSC determines the clustering
pattern $C(t)$ based on GQSI $\mathbf Q(t)$ obtained from each BS
according to \eqref{eqn:per-cluster-approximate-Q-pattern}, and
broadcasts $ C(t)$ to the active CMs of the clusters $\omega_n\in
C(t)$.
\item {\bf Step 3 [Per-cluster Power Allocation]}: Each CM $n$ ($\forall \omega_n \in C(t)$) of the
active cluster obtains CCSI $\mathbf H_{n}(t)$, CQSI $\mathbf
Q_n(t)$ and LMs $\gamma^t_{b} $ ($b \in \omega_n$) from the BSs in
its cluster, based on which,  each CM $n$ ($\forall \omega_n \in
C(t)$)  performs power allocation $\mathbf p_n(t)=\{ p_{b,k}(t): b
\in \omega_n, k\in \mathcal K\}$ according to $\Omega_{\mathbf
p_n}^*(\mathbf H_n(t), \mathbf{Q}_n(t))$.
\item {\bf Step 4 [Potential Functions Update]}: Each CM updates the {\em  per-cluster potential} $\overline V_n^{t+1}\big(\mathbf Q_n(t)\big)$ based on  CQSI $\mathbf
Q_n(t)$ according to \eqref{eqn:per-cluster-online_learn_potential}
and reports the updated potential functions to the BSC.
\item {\bf Step 5 [LMs Update]}: Each BS $b$ ($b \in \mathcal B$) calculates the total power $P_b(t)$ based on $\{p_{b,k}(t)\}$ from its CM according
to \eqref{eqn:per-BS-tx-pwr} and updates its LM $\gamma^{t+1}_{b} $
according to \eqref{eqn:per-cluster-online_learn_LM}.
\end{itemize}
\label{Alg:general_alg}
\end{Alg}
The {\em per-cluster potential} update in Step 4 and the LMs update
in Step 5 based on CCSI observation $\mathbf H_{n}(t)$ and CQSI
observation $\mathbf Q_n (t)$ at the current time slot $t$ are
further illustrated  as follows:
\begin{align}
\overline V_n^{t+1}(\mathbf Q_n^i) =&\overline V_n^t(\mathbf Q_n^i)
+\epsilon^v_{l_n^i(t)}Y_n^t(\mathbf Q_n^i)\cdot\mathbf 1[\mathbf Q_n
(t)=\mathbf Q_n^i], \ \textcolor{black}{\forall
i\in\{1,\cdots,I_{Q_n}\}}
\label{eqn:per-cluster-online_learn_potential}\\
\gamma^{t+1}_{b} = &\Gamma\Big( \gamma^{t}_{b} +
\epsilon_t^{\gamma}\big(P_b(t) - \overline P_b \big) \Big)
\label{eqn:per-cluster-online_learn_LM}
\end{align}
where $Y_n^t(\mathbf Q_n^i)= \Big(
g_{n}\big(\boldsymbol{\gamma}_n^t, \mathbf H_{n}(t),\mathbf
Q_{n}(t), p_n(t)\big)+\overline V_n^t(\mathbf Q_n(t+1)) \Big) -\Big(
g_{n}\big(\boldsymbol{\gamma}_n^t, \mathbf H_{n}(\overline t_n),\mathbf
Q_{n}^I, p_n(\overline t_n)\big)+\overline V_n^t(\mathbf Q_n(\bar t_n+1))-\bar
V_n^t(\mathbf Q_n^I)\Big)-\overline V_n^t(\mathbf Q_n^i)$,
$l_n^i(t)\triangleq\sum_{t'=0}^{t}{\mathbf 1[\mathbf Q_n(t')=\mathbf
Q_n^i, \omega_n \in  C(t')]}$ is the number of updates of $\overline V_n(\mathbf Q_n^i)$ till $t$, $\mathbf p_n(t)=\{ p_{b,k}(t): b \in
\omega_n, k\in \mathcal K\}=\Omega_{\mathbf p_n}^*(\mathbf H_n(t),
\mathbf{Q}_n(t))$, $\mathbf Q_n^I$ is the reference
state\footnote{Without lost of generality, we set reference state
$\mathbf Q_n^I=\mathbf 0$ ($\forall \omega_n$), i.e. buffer empty
for all MSs in cluster $n$, and initialize the $\overline V_n^0(\mathbf
Q_n^I)=0,\forall \omega_n$.}, $\overline t_n\triangleq\sup\{t:\mathbf
Q_n(t)=\mathbf Q_n^I, \omega_n \in C(t)\}$ is the last time slot
that the reference state $\overline V_n(\mathbf Q_n^I)$ was updated.
$\Gamma(\cdot)$ is the projection onto an interval $[0,B]$ for some
$B>0$. $\{\epsilon_{t}^v\}$ and $\{\epsilon_t^{\gamma}\}$ are the
step size sequences satisfying the following equations:
\begin{align}
 \sum_t\epsilon_t^{v}=\infty,
\epsilon_t^{v}\geq 0, \epsilon_t^{v} \rightarrow 0,\
\sum_{t}\epsilon_t^{\gamma} = \infty, \epsilon_t^{\gamma} \ge 0,
\epsilon_t^{\gamma}\to 0,\
\sum_{t}((\epsilon_t^{\gamma})^2+(\epsilon_t^v)^2)<\infty,
\frac{\epsilon_t^{\gamma}}{\epsilon_t^{v}}\to 0 \label{stepsize}
\end{align}

\subsection{Convergence Analysis for Distributive Primal-Dual Online Learning}\label{subsec:general_convergence_proof}

In this section, we shall establish the technical conditions for the
almost-sure convergence of the online distributive learning
algorithm in Algorithm \ref{Alg:general_alg}. Let $\overline{\mathbf
V}_n$ denote the $I_{Q_n}$-dimensional vector form of $\{\overline
V_n(\mathbf Q_n)\}$. For any per-cluster LM vector
$\boldsymbol{\gamma}_n$, define a vector mapping $\mathbf{T}_n:
R^{|\omega_n|}\times R^{I_{Q_n}}\rightarrow R^{I_{Q_n}}$ of cluster
$n$ with the $i$-th ($1\leq i \leq I_{Q_n}$) component mapping as:
\begin{align}
T_{n,i}(\boldsymbol{\gamma}_n,\overline{\mathbf V}_n) \triangleq
\min_{\Omega_{\mathbf p_n}(\mathbf Q_{n}^i)}\Big\{ \overline
g_{n}(\boldsymbol{\gamma}_n, \mathbf Q_{n}^i,\Omega_{\mathbf p_n}(
\mathbf Q_{n}^i))+ \sum_{\mathbf Q_{n}^j}\Pr[\mathbf Q_{n}^j|
\mathbf Q_{n}^i,\Omega_{\mathbf p_n}( \mathbf Q_{n}^i)] \overline
V_{n}(\mathbf Q_{n}^j) \Big \}\label{per-cluster-T-dis}
\end{align}
Define $\mathbf{A}^{t-1}_n\triangleq \epsilon^v_{t-1} \mathbf
P^{t}_n +(1-\epsilon^v_{t-1}) \mathbf{I}$ and
$\mathbf{B}^{t-1}_n\triangleq \epsilon^v_{t-1} \mathbf
P^{t-1}_n+(1-\epsilon^v_{t-1}) \mathbf{I}$,
where $\mathbf P^{t}_n$ is a $I_{Q_n} \times I_{Q_n}$ average
transition probability matrix for the queue of cluster $n$ with
$\Pr[\mathbf Q_{n}^j| \mathbf Q_{n}^i,\mathbf p_n^t( \mathbf
Q_{n}^i)]=\mathbb E\big[\Pr[\mathbf Q_{n}^j|\mathbf H_{n}, \mathbf
Q_{n}^i,\mathbf p_n^t$ $(\mathbf H_{n},\mathbf Q_{n}^i)]\big |
\mathbf Q_{n}^i \big] $ as its $(i,j)$-element and $\mathbf{I}$ is a
$I_{Q_n} \times I_{Q_n} $ identity matrix.

Since we have two different step size sequences $\{\epsilon^v_t\}$
and $\{\epsilon^{\gamma}_t\}$ with $\epsilon^{\gamma}_t = \mathbf
o(\epsilon^v_t)$, the {\em per-cluster potential} updates and
 the LM updates are done simultaneously but over two different timescales. During
the {\em per-cluster potential} update (timescale I), we have
$\gamma_b^{t+1}-\gamma_b^{t}=e(t)(\forall b)$, where $e(t)=\mathcal
O(\epsilon^{\gamma}_t)=\mathbf o(\epsilon^v_t)$. Therefore, the LMs
appear to be quasi-static\cite{Borkarbook:2008} during the {\em
per-cluster potential} update in
\eqref{eqn:per-cluster-online_learn_potential}. We first have the
following lemma.

\begin{Lem}(Convergence of Per-cluster Potential Learning (Time Scale I))
Assume that for every set of feasible control policies $\Omega_1$,
$\cdots$, $\Omega_{m+1}$ in the policy space, there exist a
$\delta_m=\mathcal O(\epsilon^v)>0$ and some positive integer $m$
such that
\begin{eqnarray}
[\mathbf{A}^{m}_n \cdots\mathbf{A}^{1}_n]_{i I}\geq \delta_m, \
[\mathbf{B}^{m}_n\cdots\mathbf{B}^{1}_n]_{i I}\geq \delta_m ,\ 1\le
i\le I_{Q_n}, \forall \omega_n
\label{per-cluster-promatrix-cond-dis}
\end{eqnarray}
where  $[\cdot]_{i I}$ denotes the $(i,I)$-element of the
corresponding $I_{Q_n} \times I_{Q_n}$ matrix. For stepsize sequence
$\{\epsilon_t^{v}\}$ satisfying the conditions in \eqref{stepsize},
we have $\lim_{t\rightarrow \infty}\mathbf{\overline{V}}^t_n=
\mathbf{\overline{V}}^{\infty}_n(\boldsymbol{\gamma}_n)$ a.s. ($\forall
\omega_n$) for any initial potential vector
$\mathbf{\overline{V}}^{0}_n(\boldsymbol{\gamma}_n)$ and per-cluster LM
vector $\boldsymbol{\gamma}_n$, where the steady state {\em
per-cluster potential}
$\mathbf{\widetilde{V}}^{\infty}_n(\boldsymbol{\gamma}_n)$
satisfies:
\begin{align}
\Big(T_{n,I} (\boldsymbol{\gamma}_n,\mathbf{\overline V}
^{\infty}_n(\boldsymbol{\gamma}_n) ) -{\overline{V}}^{\infty}_n(\mathbf
Q_n^I)(\boldsymbol{\gamma}_n)\Big) \mathbf e +
\mathbf{\overline{V}}^{\infty}_n(\boldsymbol{\gamma}_n)& = \mathbf T_n
(\boldsymbol{\gamma}_n, \mathbf{\overline{V}}
^{\infty}_n(\boldsymbol{\gamma}_n)), \ \forall \omega_n
\label{bellmaneqn-wp1-dis-fixed-gamma}
\end{align}
\label{Lem:convergence-dis-fixed-gamma}
\end{Lem}
\begin{proof}
Please refer to Appendix B for the proof.
\end{proof}

\begin{Rem} [Interpretation of the Conditions in Lemma \ref{Lem:convergence-dis-fixed-gamma}]
Note that $\mathbf A_n^t$ and $\mathbf B_n^t$ are related to an
equivalent transition matrix of the underlying Markov chain.
Condition in \eqref{per-cluster-promatrix-cond-dis} simply means
that state $\mathbf Q_n^I$ is accessible from all the $\mathbf
Q_n^i$ states after some finite number of transition steps. This is
a very mild condition and will be satisfied in most of the cases we
are interested.  ~ \hfill\QED
\end{Rem}

 On the other
hand, during the LM update (timescale II), we have
$\lim_{t\to\infty}\|\mathbf{\overline V}_n^t-\mathbf{\overline
V}_n^{\infty}(\boldsymbol{\gamma}_n^t)\|=0$ w.p.1. by the corollary
2.1 of \cite{Borkartwotimescales:1997}. Hence, during the LM update
in \eqref{eqn:online_learn_LM}, the {\em  per-cluster potential} is
seen as almost equilibrated. The convergence of the LM is summarized
below.

\textcolor{black}{
\begin{Lem}(Convergence of LM over Timescale II): For the same conditions in Lemma \ref{Lem:convergence-dis-fixed-gamma}, we have $\lim_{t\to \infty} \boldsymbol
\gamma^t=\boldsymbol \gamma^{\infty}$ a.s. where $\boldsymbol
\gamma^{\infty}$ satisfies the power constraints of all the BSs in
\eqref{eqn:tx-pwr-constraint}. 
\label{Lem:convergence-LM}
~ \hfill\QED
\end{Lem}
}
\begin{proof}
Please refer to Appendix C for the proof.
\end{proof}

Based on the above lemmas, we can summarize the convergence
performance of the online {\em per-cluster potential} and LM
learning algorithm in the following theorem:
\begin{Thm}(Convergence of Online Learning Algorithm)
 For the same conditions as
in Lemma \ref{Lem:convergence-dis-fixed-gamma}, we have $(\mathbf
{\overline V}_n^{t},\boldsymbol{\gamma}_n^t)\to(\mathbf{\overline
V}_n^{\infty},\boldsymbol{\gamma}_n^{\infty})$ a.s. ($\forall
\omega_n, b\in \omega_n, k\in \mathcal K$), where $\mathbf{\overline
V}_n^{\infty}$ and $\boldsymbol{\gamma}_n^{\infty}$ satisfy
\begin{align}
 \Big(T_{n,I}
(\boldsymbol{\gamma}_n ^{\infty}, \mathbf{\overline{V}} ^{\infty}_n )
-{\overline{V}}^{\infty}_n(\mathbf Q_n^I)\Big) \mathbf e +&
\mathbf{\overline{V}}^{\infty}_n = \mathbf T_n
(\boldsymbol{\gamma}_n^{\infty}, \mathbf{\overline{V}} ^{\infty}_n), \
\forall \omega_n \label{bellmaneqn-wp1-dis-converged-gamma}
\end{align}
and the power constraints of all the BSs in
\eqref{eqn:tx-pwr-constraint}. ~ \hfill\QED
\label{Thm:per-cluster-convergence-online-learning}
\end{Thm}

\section{Application to Network MIMO Systems with Poisson Arrival}\label{sec:application_poisson_exponential}

In this section, we shall illustrate  the online primal-dual
distributive learning algorithm  for network MIMO systems under
Poisson packet arrival and exponential  packet size distribution.


\subsection{Dynamics of System State under Poisson Packet Arrival and Exponential Distributed
Packet Size}

Under Poisson assumption, we could consider packet flow rather than
bit flow. Specifically, let $\mathbf{A}(t)=\{A_{b,k}(t):
b\in\mathcal{B}, k\in \mathcal{K}\}$ and
$\mathbf{N}(t)=\{N_{b,k}(t): b\in\mathcal{B}, k\in \mathcal{K}\}$ be
the random new packet arrivals and the corresponding packet sizes
for the $BK$ users in the multicell network at the end of the $t$-th
scheduling slot, respectively. $\mathbf Q(t)$ and $N_Q$ denotes the
GQSI matrix (number of packets) and maximum buffer size (number of
packets).

\begin{Asump}[Poisson Source Model]
The packet arrival process $A_{b,k}(t)$ is i.i.d. over scheduling
slots following Poisson distribution with average arrival rate
$\mathbb{E}[A_{b,k}]=\lambda_{b,k}$, and independent w.r.t.
$\{(b,k)\}$. The random packet size $N_{b,k}(t)$ is i.i.d. over
scheduling slots following an exponential distribution with mean
packet size $\overline N_{b,k}$, and independent w.r.t. $\{(b,k)\}$. ~
\hfill\QED \label{Asump:A-N}
\end{Asump}

Given a stationary policy, define the conditional mean departure
rate of packets of MS $(b,k)$ at the $t$-th slot (conditioned on
$\boldsymbol{\chi}(t)$) as
$\mu_{b,k}\big(\boldsymbol{\chi}(t)\big)=\mathbb{E}[R_{b,k}(\boldsymbol{\chi}(t))/
N_{b,k}|\boldsymbol{\chi}(t)]=R_{b,k}\big(\boldsymbol{\chi}(t)\big)/
\overline N_{b,k}$.

\begin{Asump}[Time Scale Separation]
The slot duration $\tau$ is sufficiently small compared with the
average packet interarrival time as well as conditional average
packet service time\footnote{This assumption is reasonable in
practical systems, such as WiMax. 
In practical systems, an application level packet may have mean
packet length spanning over many time slots (frames) and this
assumption is also adopted in a lot of literature such as
 \cite{Sadiq:2009,Baris:2009}.}, i.e. $\lambda_{b,k}\tau \ll 1$ and
$\mu_{b,k}\big(\boldsymbol{\chi}(t)\big) \tau \ll 1$. ~ \hfill\QED
\label{Asump:tau}
\end{Asump}


By the memoryless property of the exponential distribution, the
remaining packet length (also denoted as $\mathbf N(t)$) at any slot
$t$ is also exponential distributed. Given a stationary control
policy $\Omega$, the conditional probability (conditioned on
$\boldsymbol{\chi}(t)$) of a packet departure event at the $t$-th
slot is given by
\begin{align}
\Pr \Big[ \frac{N_{b,k}(t)}{R_{b,k}(t)} < \tau |
\boldsymbol{\chi}(t),\Omega(\boldsymbol{\chi}(t)) \Big] = \Pr
\Big[\frac{N_{b,k}(t)}{\overline N_{b,k}} <
\mu_{b,k}(\boldsymbol{\chi}(t))\tau \Big]= 1 - \exp
(-\mu_{b,k}(\boldsymbol{\chi}(t))\tau) \approx
\mu_{b,k}(\boldsymbol{\chi}(t)) \tau \nonumber
\end{align}
where the last equality is due to Assumption \ref{Asump:tau}. Note
that under Assumption \ref{Asump:tau}, the probability for
simultaneous arrival, departure of two or more packets from the same
queue or different queues and simultaneous arrival as well as
departure in a slot are $\mathcal{O}\big(\lambda_{b,k} \tau \cdot
\lambda_{b',k'} \tau \big)$,
$\mathcal{O}\big((\mu_{b,k}(\boldsymbol{\chi}(t)) \tau \cdot
(\mu_{b',k'}(\boldsymbol{\chi}(t)) \tau \big)$ and
$\mathcal{O}\big(\lambda_{b,k}
\tau\cdot\mu_{b,k}(\boldsymbol{\chi}(t)) \tau\big)$ respectively,
which are asymptotically negligible. 
Therefore, the transition kernel of the QSI evolution in this
example can be simplified as: \textcolor{black}{
\begin{align}
&\Pr[\mathbf Q_{n}'|\mathbf Q_{n},\Omega_{\mathbf p_n}(\mathbf
Q_{n})]=  \left\{
\begin{array}{ll}
\lambda_{b,k}\tau &\text{if }\mathbf Q_{n}'=\mathbf Q_{n}+ \mathbf e_{b,k}\\
\mathbb E[\mu_{b,k}(\mathbf H_{n},\mathbf Q_{n}) | \mathbf Q_{n}]\tau  & \text{if }\mathbf Q_{n}'=\mathbf Q_{n}- \mathbf e_{b,k}\\
1-\sum_{b\in\omega_n}\sum_{k\in \mathcal K} \big(\mathbb
E[\mu_{b,k}(\mathbf H_{n},\mathbf Q_{n}) | \mathbf
Q_{n}]+\lambda_{b,k}\big) \tau & \text{if }\mathbf Q_{n}'=\mathbf
Q_{n}
\end{array}
\right.\label{eqn:per-cluster-tran-prob}
\end{align}
} where \textcolor{black}{$\mu_{b,k}(\mathbf H_{n},\mathbf
Q_{n})=R_{b,k}(\mathbf
H_{n},\mathbf Q_{n})\tau/\overline N_{b,k}$}  
and $\mathbf{e}_{b,k}$ denotes the $|\omega_n|\times K$ matrix with
element 1 corresponding to MS $(b,k)$ and all other elements 0.

%

\subsection{Decomposition of the Per-cluster Potential Function}

Observe that the cardinality of the per-cluster system state $
I_{Q_n}= (N_Q+1)^{|\omega_n|K}$ is still exponential in the number
of all the MSs in cluster $n$, i.e. $|\omega_n|K$. In the following
lemma, we shall show that the {\em per-cluster potential} can be
further decomposed into {\em per-cluster per-user potential}, which
leads to linear order of growth in the cardinality of the state
space, i.e. \textcolor{black}{ $\sum_n |\omega_n|K (N_Q+1)$}.

\begin{Lem} [Decomposition of Per-cluster Potential]
The {\em per-cluster potential} $\overline V_{n}(\mathbf Q_{n})$
($\forall \omega_n$) defined by the fixed point equation in
\eqref{eqn:bellman-per-cluster-potential}  can be decomposed into
the sum of the {\em per-cluster per-user potential} functions
$\{\overline V_{n,(b,k)}(Q)\}$, i.e. $\overline V_{n}(\mathbf Q_{n})
= \sum_{b\in\omega_n,k\in\mathcal K}\overline V_{n,(b,k)}(Q_{b,k})$,
where $\{\overline V_{n,(b,k)}(Q )\}$ satisfy the following {\em
per-cluster
  per-user potential fixed point equation}: ($\forall Q\in\mathcal Q$)
\textcolor{black}{
\begin{align}
\theta_{n,(b,k)} + \overline V_{n,(b,k)}(Q) = \min_{\Omega_{
p_{b,k}}(Q)} \Big \{ \overline
g_{n,(b,k)}(\boldsymbol{\gamma}_n,Q,\Omega_{ p_{b,k}}(Q)) +
\sum_{Q'}&\Pr[Q'| Q,\Omega_{ p_{b,k}}(Q)]\overline
V_{n,(b,k)}(Q')\Big\}\label{eqn:bellman-per-cluster-per-user-potential}
\end{align}
\begin{align}
\text{where }\ \ \overline g_{n,(b,k)}(\boldsymbol{\gamma}_n,Q,\Omega_{
p_{b,k}}(Q)) =&\mathbb E\underbrace{\Big[\beta_{b,k}f(Q)+
p_{b,k}(\mathbf H_{n},Q)\sum_{b'\in\omega_n}\gamma_{b'}\|\mathbf{
w}_{(b,k),b'}\|^2 \Big |
Q\Big]}_{g_{n,(b,k)}(\boldsymbol{\gamma}_n,\mathbf H_{n},Q,
p_{b,k}(\mathbf H_{n},Q))}
\label{eqn:per-cluster-per-user-per-stage-cost}
\end{align}
\begin{align}
& \Pr[Q'| Q,\Omega_{ p_{b,k}}(Q)]=\mathbb E\big[\Pr[Q'|
\mathbf H_{n},Q,\Omega_{ p_{b,k}}(\mathbf H_{n},Q)]\big |
Q\big]\nonumber\\
=& \left\{
\begin{array}{ll}
\lambda_{b,k}\tau &\text{if }Q'=Q+ 1\\
\mathbb E\big[\mu_{b,k}(\mathbf H_{n},Q)\big |
Q\big]\tau & \text{if }Q'=Q- 1\\
1-\big(\mathbb E\big[\mu_{b,k}(\mathbf H_{n},Q)\big |
Q\big]+\lambda_{b,k}\big) \tau & \text{if }Q'=Q
\end{array}
\right.\label{eqn:per-cluster-per-user-tran-prob}
\end{align}
} where \textcolor{black}{$\mu_{b,k}(\mathbf
H_{n},Q)=R_{b,k}(\mathbf
H_{n},Q)\tau/\overline N_{b,k}$}. 
\label{Lem:per-cluster-per-user-conditional-potential} ~ \hfill\QED
\end{Lem}
\begin{proof}
Please refer to Appendix D for the proof.
\end{proof}


\subsection{Per-Stage QSI-aware Interference Game for Power Allocation at the
CMs}\label{subsec:per-stage-game}


In order to determine the power control action (Step 3 in Algorithm
\ref{Alg:general_alg}) in a distributive manner at each CM, we shall
formulate the power allocation of each cluster $n$ ($\omega_n\in C$)
as a non-cooperative game\cite{PalomarmuMIMOunifiedview:2008}. The
players are the CMs, and the payoff function for each cluster $n$
($\omega_n\in C$) is defined as \textcolor{black}{
\begin{align}
R_n(\mathbf p_n,\mathbf p_{-n})=g_{n}(\boldsymbol{\gamma}_n, \mathbf
H_{n},\mathbf Q_{n},\Omega_{p_n}(\mathbf H_{n},\mathbf Q_{n}))+
\sum_{\mathbf Q_{n}'}\Pr[\mathbf Q_{n}'|\mathbf H_{n},\mathbf
Q_{n},\Omega_{p_n}(\mathbf H_{n},\mathbf Q_{n})] \overline
V_{n}(\mathbf Q_{n}')\nonumber
\end{align}
}
Each CM is a player in the game specified by:
\begin{align}
(\mathcal G):\quad \textcolor{black}{\min_{\mathbf p_n} R_n(\mathbf
p_n,\mathbf p_{-n}),\ \forall \omega_n\in C}
\label{eqn:per-cluster-game}
\end{align}
where $\mathbf p_n=\{p_{b,k}:b \in \omega_n, k\in \mathcal K\}$ is
power allocation of cluster $n$ ($\omega_n \in C$) and  $\mathbf
p_{-n}=\cup_{\omega_{n'} \in C, n' \neq n}\mathbf p_n$ is the power
allocation of all other clusters indirectly observed through
interference measure from MSs in cluster $n$.
It can be shown that the solution of the game $\mathcal G$ in
\eqref{eqn:per-cluster-game}, i.e. Nash Equilibrium (NE)  can be
characterized by the following fixed-point equation:
\begin{align}
p_{b,k}^* = WF_{n,(b,k)}(\mathbf p^*_{-n}),\quad \forall
b\in\omega_n, k\in\mathcal K, \omega_n\in C
\label{eqn:NE-fixed-point}
\end{align}
where the waterfilling operator $WF_{n,(b,k)}(\cdot)$ ($\forall b
\in \omega_n, k\in \mathcal K$) is defined as:
\begin{align}
WF_{n,(b,k)}(\mathbf p_{-n})= \Big(\frac{ \frac{\tau}{\overline
N_{b,k}}\Delta \overline V_{n,(b,k)}(Q_{b,k})}
{\sum_{b'\in\omega_n}\gamma_{b'}\parallel \mathbf{
w}_{(b,k),b'}\parallel^2} - (1+I_{n,(b,k)})
\Big)^+\label{eqn:per-cluster-per-user-pwr-allo-with-intf}
\end{align}
where $I_{n,(b,k)}= \sum_{\substack{\omega_{n'}\in C\\ n'\neq
n}}\sum_{\substack{b''\in\omega_{n'}\\ k''\in\mathcal K }}
(\sum_{b'\in\omega_{n'}} \big|\mathbf{h}_{(b,k),b'}\mathbf
w_{(b'',k''),b'}\big|^2) p_{b'',k''}$ is the inter-cell interference
measured by the MS ($b,k$) in cluster $n$ ($\forall b \in \omega_n,
k \in \mathcal K, \omega_n \in C$).
For notation convenience, let the $BK\times 1$ vector $\mathbf p$
and $\mathbf{WF}$ denote the vector form of $p_{b,k}$ and
$WF_{n,(b,k)}$ ($\forall b \in \omega_n, k\in \mathcal K, \omega_n
\in C$), respectively.

We propose a QSI-aware Simultaneous Iterative Water-filling
Algorihtm (QSIWFA) to  achieve the NE of the game $\mathcal G$
distributively. At each iteration, given the measurement  of
interference generated by other clusters in the previous iteration,
the overall power allocations are updated by the active CMs
simultaneously according to
\begin{align}
\mathbf p^{\nu+1} = \mathbf{WF}(\mathbf
p^{\nu})\label{eqn:IWFA-vector}
\end{align}

\begin{Rem}[Multi-level Water-filling Structure of QSIWFA]
The waterfilling operator $WF_{n,(b,k)}(\cdot)$  in
\eqref{eqn:per-cluster-per-user-pwr-allo-with-intf}  is function of
both CCSI and CQSI. It has the form of {\em multi-level
water-filling} where the power is allocated according to the CCSI in
terms of $\mathbf{w}_{(b,k),b'}$ but the water-level is adaptive to
the CQSI (indirectly via $\Delta \overline V_{n,(b,k)}(Q_{b,k})$). ~
\hfill\QED
\end{Rem}

Next, we shall discuss the existence, uniqueness of the NE of game
$\mathcal G$ and convergence of the multi-level QSIWFA in
\eqref{eqn:IWFA-vector}. Define the $BK\times BK$ matrix $\mathbf S$
with its element $[\mathbf S]_{(b,k),(b',k')}$ given by:
\begin{align}\label{eqn:S-linear-mapping-waterfilling}
[\mathbf S]_{(b,k),(b',k')}=\left\{
\begin{array}{ll}
\sum_{b''\in\omega_{n'}}
 \big|\mathbf{h}_{(b,k),b''}\mathbf w_{(b',k'),b''}\big|^2  & \text{if}\ n'\neq n\\
0 &\text{if}\ n'=n
\end{array}
\right. \forall b\in\omega_n, b'\in\omega_{n'}, k,k'\in\mathcal K
\end{align}
Given some $BK \times 1$ vector $\mathbf u$ with each component
positive, let $\|\cdot \|_{\infty,\text{vec}}^{\mathbf u}$ and
$\|\cdot\|_{\infty,\text{mat}}^{\mathbf u}$ denote the vector
weighted maximum norm and the matrix norm defined in
\cite{PalomarmuMIMOunifiedview:2008}, separately. Then, we have
$\|\mathbf{WF}(\mathbf p)\|_{\infty,\text{vec}}^{\mathbf
u}\triangleq\max_{b,k}\frac{(WF_{b,k}(\mathbf p))^2}{u_{b,k}}$ and
$\|\mathbf S\|_{\infty,\text{mat}}^{\mathbf
u}\triangleq\max_{b,k}\frac{1}{u_{b,k}}\sum_{b'\in\mathcal
B,k'\in\mathcal K}[\mathbf S]_{(b,k),(b',k')}u_{b',k'}$, where
$\mathbf S\in\mathbb R^{BK\times BK}$, and $[\cdot]_{(b,k),(b',k')}$
denotes the element in the row corresponding to MS ($b,k$) and the
column corresponding to MS ($b',k'$). The convergence of the QSIWFA
in \eqref{eqn:IWFA-vector} is summarized in the following lemma.

\begin{Lem}(Convergence of the QSIWFA)\label{Lem:convergence-IWFA}
If $\|\mathbf S\|^{\mathbf u}_{\infty,\text{mat}}<1$ is satisfied
for some $\mathbf u>0$, then the mapping $\mathbf{WF}$ is a
contraction mapping with modulus $\alpha = \|\mathbf S\|^{\mathbf
u}_{\infty,\text{mat}}$ w.r.t the norm
$\|\cdot\|_{\infty,\text{vec}}^{\mathbf u}$. The NE of game
$\mathcal G$ exists and is unique. As $\nu\rightarrow \infty$. The
QSIWFA in \eqref{eqn:IWFA-vector} converges to the unique NE of game
$\mathcal G$ which is the solution to the fixed point equation in
\eqref{eqn:NE-fixed-point}. ~ \hfill\QED
\end{Lem}
\begin{proof}
Please refer to Appendix E for the proof.
\end{proof}

\begin{figure}[h]
  \begin{center}
  \includegraphics[height=6cm, width=8cm]{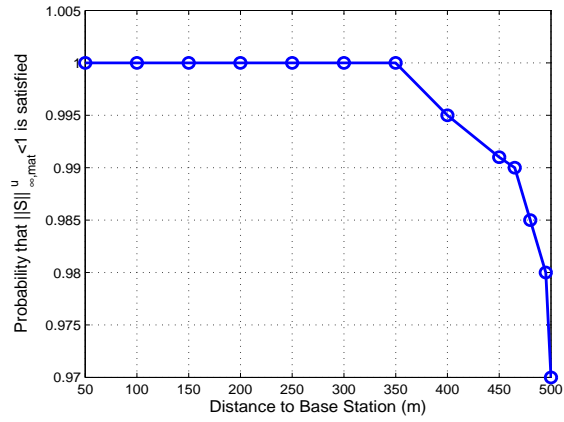}
  \caption{\textcolor{black}{Probability that condition $\|\mathbf S\|^{\mathbf u}_{\infty,\text{mat}}<1$ is satisfied versus user
  location. The number of cells $B=19$, the number of MSs per cell $K=1$, the
maximum cluster size $N_B=3$, the number of transmit antennas
$N_t=4$.}} \label{Fig:prob_sat}
  \end{center}
  \end{figure}

\textcolor{black}{
\begin{Rem} [Interpretation of Sufficient Condition for QSIWFA Convergence]
The intuitive meaning for the condition $\|\mathbf S\|^{\mathbf
u}_{\infty,\text{mat}}<1$ is that the inter-cluster interference is
sufficiently small compared with the signal power from cooperative
BSs in the same cluster\cite{PalomarmuMIMOunifiedview:2008}. This
happens with high probability because the interference from the
inter-cluster BS has been reasonably attenuated due to the geometry
of the cluster topology. Fig. \ref{Fig:prob_sat} also illustrates
that the condition $\|\mathbf S\|^{\mathbf u}_{\infty,\text{mat}}<1$
can be satisfied with high probability.
\end{Rem}
}

\subsection{Compact Queue State in Online Primal-Dual Distributive Learning Algorithm}


To further reduce the memory size as well as the frequency of
clustering updates at the BSC, we shall use the feature-based linear
architecture \cite{featurebased:1996} to approximate the original
{\em per-cluster per-user potential} functions. Specifically, we
define the following {\em compact queue state}.

\begin{Def}[Compact Queue State] Define the {\em compact
queue state}  as $Q=qd$  ($q=0,\cdots,l_q=\lfloor
\frac{N_Q}{d}\rfloor\}$), where $d$ ($d\leq N_Q, d \in \mathcal N
^+$) is the corresponding resolution level. The approximate
potential functions of the compact queue states $\{\widetilde
V_{n,(b,k)}(q): q=0,\cdots,l_q\}$ are defined as {\em compact
per-cluster per-user potential} functions. ~ \hfill\QED
\end{Def}
Therefore, the linear approximation of the
original {\em per-cluster per-user potential} functions is given by
\begin{align}
\overline V_{n,(b,k)}(qd+l)= \widetilde V_{n,(b,k)}(q) +
\frac{l}{d}\Big(\widetilde V_{n,(b,k)}(q+1)-\widetilde
V_{n,(b,k)}(q)\Big)\nonumber\\
\forall l\in \{0,\cdots, d-1\}, \quad q\in \{0,\cdots, l_q-1\}
\label{eqn:interpolation}
\end{align}
Let $\overline {\mathbf V}_{n,(b,k)}\triangleq(\overline
{V}_{n,(b,k)}(0),\cdots,\overline {V}_{n,(b,k)}(N_Q))^T$ and $\widetilde
{\mathbf V}_{n,(b,k)}\triangleq(\widetilde
{V}_{n,(b,k)}(0),\cdots,\widetilde {V}_{n,(b,k)}(l_q))^T$ be the
vector form of the {\em per-cluster per-user potential} functions
and the {\em compact per-cluster per-user potential} functions of MS
($b,k$) in cluster $\omega_n$, respectively. Accordingly, their
relationship is given by $\mathbf{\overline V}_{n,(b,k)}=\mathbf M
\mathbf{\widetilde V}_{n,(b,k)}$ and $\mathbf{\widetilde
V}_{n,(b,k)}=\mathbf M^{\dag} \mathbf{\overline V}_{n,(b,k)}$, where
$\mathbf M$ is the $(N_Q+1)\times (l_q+1)$ matrix with
$(qd+l,q)$-element $\frac{d-l}{d}$, $(qd+l,q+1)$-element
$\frac{l}{d}$, $(l_qd,l_q)$-element 1 ($\forall q\in\{0,l_q-1\}$,
$\forall l\in\{0,d-1\}$) and all other elements 0, and $\mathbf
M^{\dag}$ is $(l_q+1)\times (N_Q+1)$ matrix with $(q,qd)$-element
($\forall q\in\{0,l_q\}$) 1 and all other elements 0\footnote{Note
that when $d=1$, $\mathbf M$ and $\mathbf M^{\dag}$ become
$(N_Q+1)\times (N_Q+1)$ identity matrix, and the feature state space
is equivalent to the original state space. In other words,  $\mathbf
{\widetilde V}_{n,(b,k)}=\mathbf{\overline V}_{n,(b,k)}$. $\mathbf
{\widetilde V}_{n,(b,k)}$ can be obtained by online learning via
stochastic approximation.}.

Applying Algorithm \ref{Alg:general_alg} to estimate the {\em
compact per-cluster per-user potential} functions and LMs with {\em
per-stage QSI-aware interference game} in Section
\ref{subsec:per-stage-game} for power allocation, we obtain the
distributive online learning algorithm for Poisson arrival and
exponential packet size distribution  as  illustrated in Fig.
\ref{fig:system-flow}. Specifically, the {\em compact per-cluster
per-user potential} update and LMs update based on CCSI observation
$\mathbf H_{n}(t)$ and CQSI observation $\{Q_{n,(b,k)}(t): \forall
b\in \omega_n, k \in \mathcal K\}$) are given by:
\begin{align}
\widetilde V_{n,(b,k)}^{t+1}(q) =&\widetilde V_{n,(b,k)}^{t}(q)
+\epsilon^v_{l_{n,(b,k)}^q(t)}Y_{n,(b,k)}^t(q)\cdot\mathbf
1[Q_{b,k}(t)=qd, \omega_n \in C(t)]
\label{eqn:online_learn_potential}\\
 Y_{n,(b,k)}^t(q)=&  \Big(
g_{n,(b,k)}(\boldsymbol{\gamma}_n^t,\mathbf H_{n}(t),Q_{b,k}(t),
p_{b,k}(t)) +\mu_{b,k}\tau\frac {\widetilde
V_{n,(b,k)}^{t}((q-1)^+)-\widetilde V_{n,(b,k)}^{t}(q)}{d}\nonumber
\\
 & +\mathbf {1}[A_{n,(b,k)}(t)]\frac {\widetilde
V_{n,(b,k)}^{t}((q+1)\wedge l_q) -\widetilde V_{n,(b,k)}^{t}(q)}{d}
\Big)-\Big(   g_{n,(b,k)}(\boldsymbol\gamma^t_n, \mathbf H_{n}(\overline
t),
Q^I, p_{b,k}(\overline t))\nonumber \\
& +\mathbf {1}[A_{n,(b,k)}(\overline t)]\frac {\widetilde
V_{n,(b,k)}^{t}(q^I+1)-\widetilde V_{n,(b,k)}^{t}(q^I)}{d}
\Big)\nonumber
\\
\gamma^{t+1}_{b} = &\Gamma\Big( \gamma^{t}_{b} +
\epsilon_t^{\gamma}\big(P_b^*(t) - \overline P_b \big) \Big)
\label{eqn:online_learn_LM}
\end{align}
where $l_{n,(b,k)}^q(t)\triangleq\sum_{t'=0}^{t}{\mathbf
1[Q_{b,k}(t')=qd, \omega_n \in  C(t')]}$ is the number of updates of
$\widetilde V_{n,(b,k)}(q)$ till $t$, $
p_{b,k}(t)=\Big(\frac{\frac{\tau}{\overline {N}_{b,k}}\Delta \overline
V_{n,(b,k)}^t\big(Q_{b,k}(t)\big)}
{\sum_{b'\in\omega_n}\gamma_{b'}\parallel \mathbf{
w}_{(b,k),b'}(t)\parallel^2} -1\Big)^+$,
$A_{n,(b,k)}(t)\triangleq\{Q_{b,k}(t+1) = Q_{b,k}(t)+1, \omega_n \in
 C(t)\}$ is the arrival event, $q_I$ is the reference
state\footnote{Without lost of generality, we set reference state
$q^I=Q^I=0$, and initialize the $\widetilde
V_{n,(b,k)}^0(q^I)=0,\forall n,b,k$.}, $\overline
t\triangleq\sup\{t:Q_{b,k}(t)=Q^I, \omega_n \in  C(t)\}$ is the last
time slot that the reference state $\widetilde V_{n,(b,k)}(q^I)$ was
updated, $P_b^*(t) = \sum_{b'\in\omega_n}\sum_{k\in\mathcal K}
\|\mathbf{ w}_{(b',k),b}\|^2\cdot p_{b',k}^*(t)$ is the total
transmit power of BS $b$ determined by the per-stage QSI-aware
interference game.

\section{Simulation Results and Discussions}\label{sec:numerical_result}
In this section, we shall compare the proposed two-timescale
delay-optimal dynamic clustering and power allocation design with
three baselines. Baseline 1 refers to the Fixed Channel Assignment
(FCA) without cooperation among BSs in standard cellular systems
with frequency reuse factor (FRF) 7. Baseline 2 and Baseline 3 refer
to {\em static clustering} and {\em greedy dynamic clustering}
\cite{Howard2007:networkmimo,RobertHeath2009:networkmimo,Belllab2008:networkmimoclustering,Gesbert2008:dynamicclustering}.
For any given clustering pattern, optimal power allocation is
performed at each BS (Baseline 1) or CM (Baseline 2 and Baseline 3)
based on available instantaneous CSI to maximize sum throughput of
the cluster. In the simulation, we consider a cellular system with
19 BSs, each has a coverage of 500m. We apply the Urban Macrocell
Model in 3GPP with path loss model given by
$PL=34.5+35\log_{10}(r)$, where $r$ (in m) is the distance from the
transmitter to the receiver. Each element of the small scale fading
channel matrix is $\mathcal{CN}(0,1)$ distributed. The total BW is
10MHz. We consider Poisson arrival with average arrival rate
$\lambda_{b,k}$ (pck/slot). The scheduling slot duration $\tau$ is
5ms. The maximum buffer size $N_Q$ is 9 and the mean packet size
$\overline N_{b,k}=264$ Kbyte.

\begin{figure}[h]
  \begin{center}
    \subfigure[\scriptsize{\textcolor{black}{Average delay per user versus transmit power at the number
  of  antenna per-BS $N_t=4,2$ and the maximum cluster size $N_B=3$.}}]
    {\resizebox*{8cm}{6cm}{\includegraphics{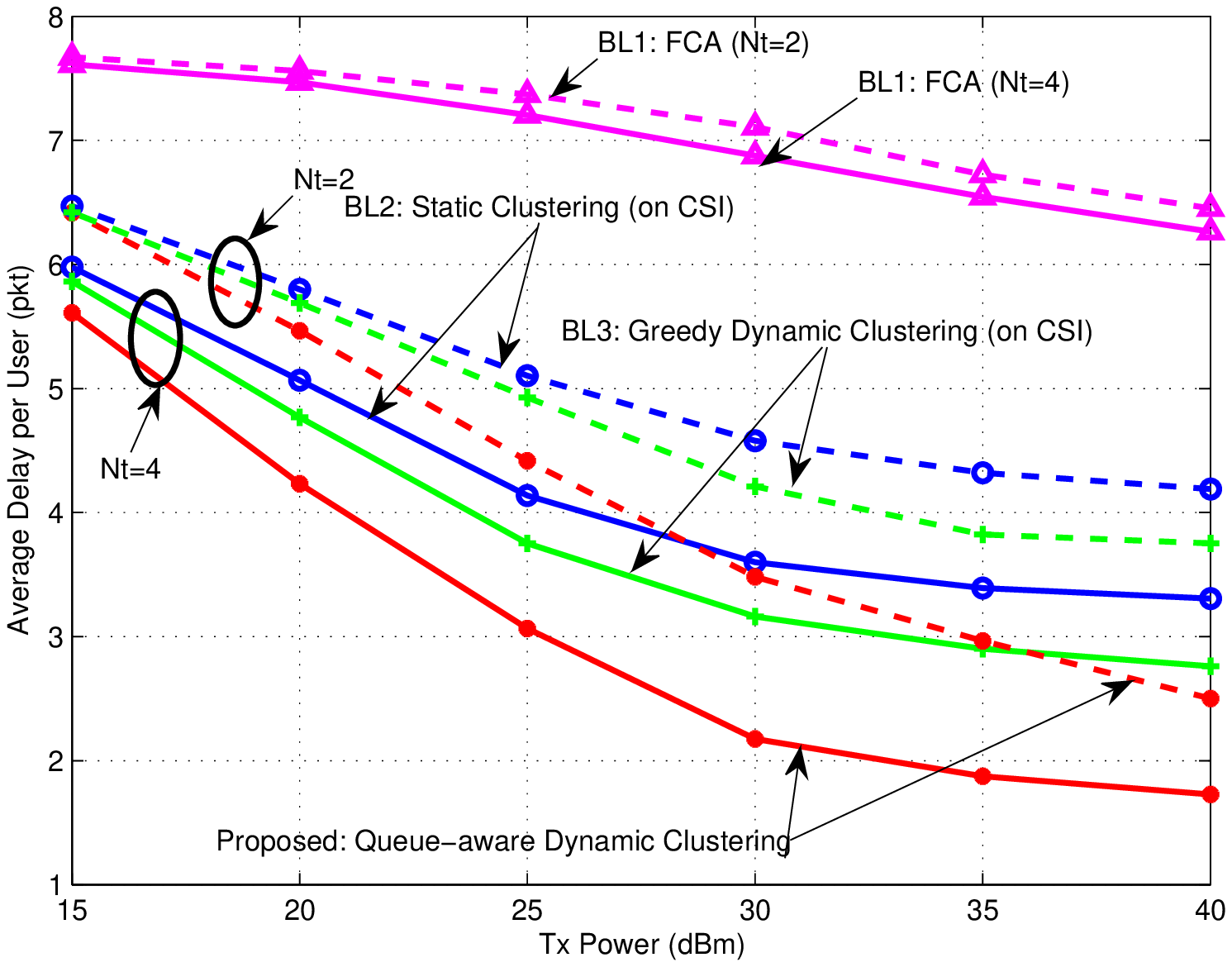}}}
    \subfigure[\scriptsize{Average delay per user versus transmit power at the maximum cluster size $N_B=3,4$ and the number
  of  antenna per-BS $N_t=4$.}]
    {\resizebox*{8cm}{6cm}{\includegraphics{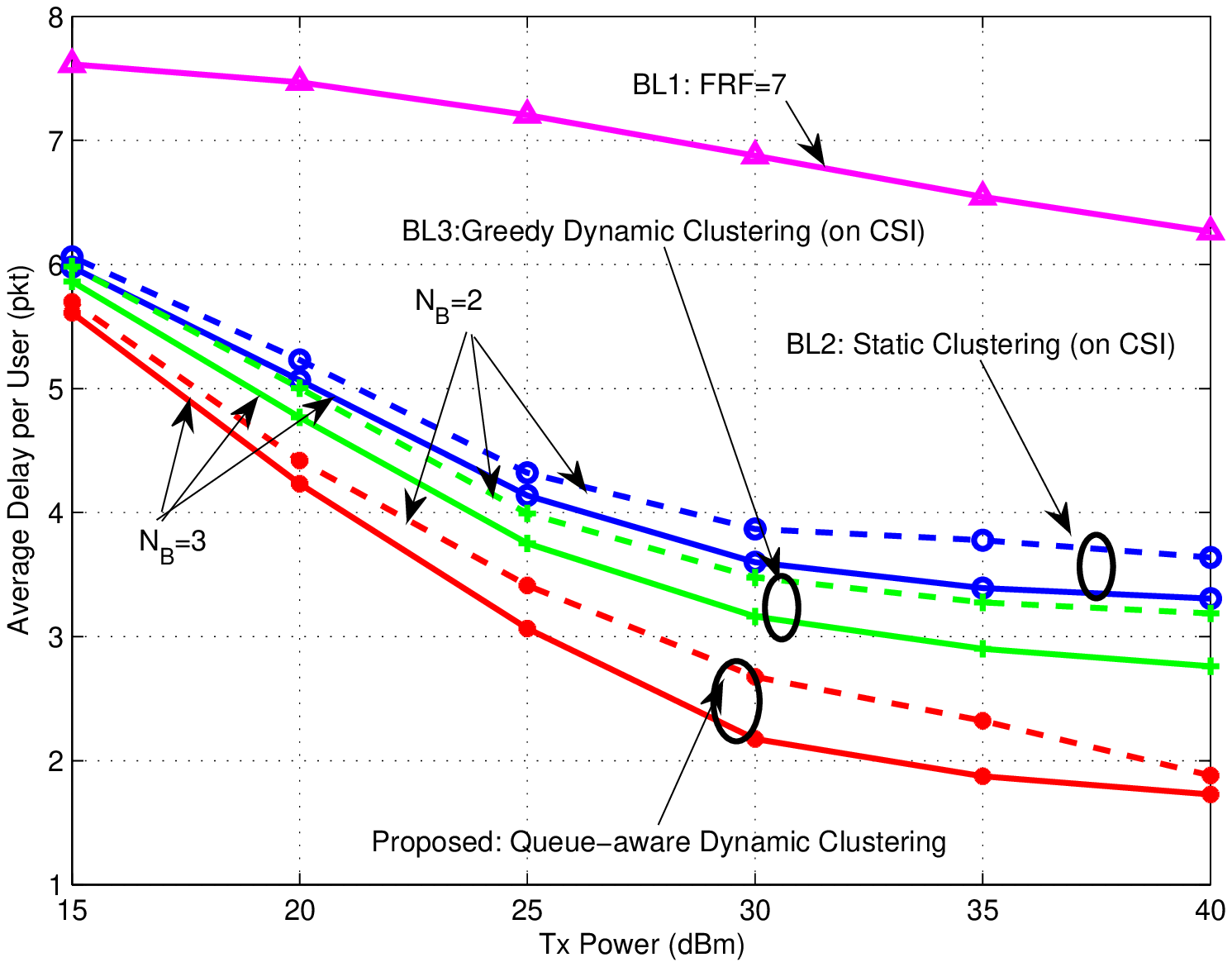}}}
    \end{center}
      \caption{Average delay per user versus transmit power. The number of MSs per BS $K=1$, the average
  arrival rate $\lambda_{b,k}=10$ pck/slot and the resolution level
  $d=3$.}
      \label{Fig:antenna_clustersize}
  \end{figure}

Fig. \ref{Fig:antenna_clustersize} (a) illustrates the average delay
per user versus transmit power with $N_t=4$ and $N_t=2$ and the
maximum cluster size $N_B =3$. The average delay of all the schemes
decreases as the transmit power or the number transmit antenna
increases. Observe that the performance of Baseline 1 is inferior to
that of Baseline 2 and 3, and this illustrates the gain behind base
station cooperations in network MIMO systems. Furthermore, Baseline
3 outperforms Baseline 2, illustrating the benefit of dynamic
clustering in network MIMO. Finally, there is significant
performance gain of the proposed scheme compared to all baselines.
This gain is contributed by the QSI-aware dynamic clustering as well
as the {\em QSIWFA} for power control. Fig.
\ref{Fig:antenna_clustersize} (b) illustrates the average delay
versus average transmit power under different maximum cluster size
$N_B=3$ and $N_B=2$. Similar observations about the performance gain
could be made. \textcolor{black}{Table \ref{tab:complexity}
illustrates the complexity in terms of the CPU time of the baselines
and the proposed scheme. It can be seen that the the proposed scheme
can achieve significant performance gain with reasonable complexity
compared with the complexity of the baselines.}

\begin{figure}[t]
  \begin{center}
  \includegraphics[height=6cm, width=8cm]{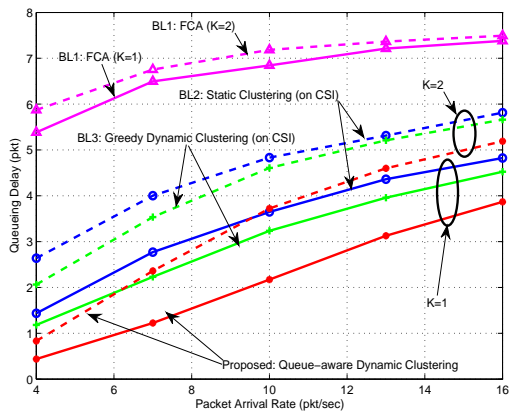}
  \caption{Average delay per user versus per-user loading (average
  arrival rate $\lambda_{b,k}$) at the number of MSs per BS $K=1$ and
  $K=2$ at the transmit power $\bar P_b=35$ dbm. The maximum cluster
  size $N_B=3$, the number of transmit antenna at each BS $N_t=4$, and
  the resolution level $d=3$.} \label{Fig:sdmaloading}
  \end{center}
  \end{figure}

Fig. \ref{Fig:sdmaloading} illustrates the average delay versus per
user loading (average arrival rate $\lambda_{b,k}$) at transmit
power of $\overline P_b=30$ dbm and number of MS per BS $K=1,2$. The
proposed scheme achieved significant gain over all the baselines
across a wide range of input loading.

\begin{figure}[t]
  \begin{center}
  \includegraphics[height=6cm, width=8cm]{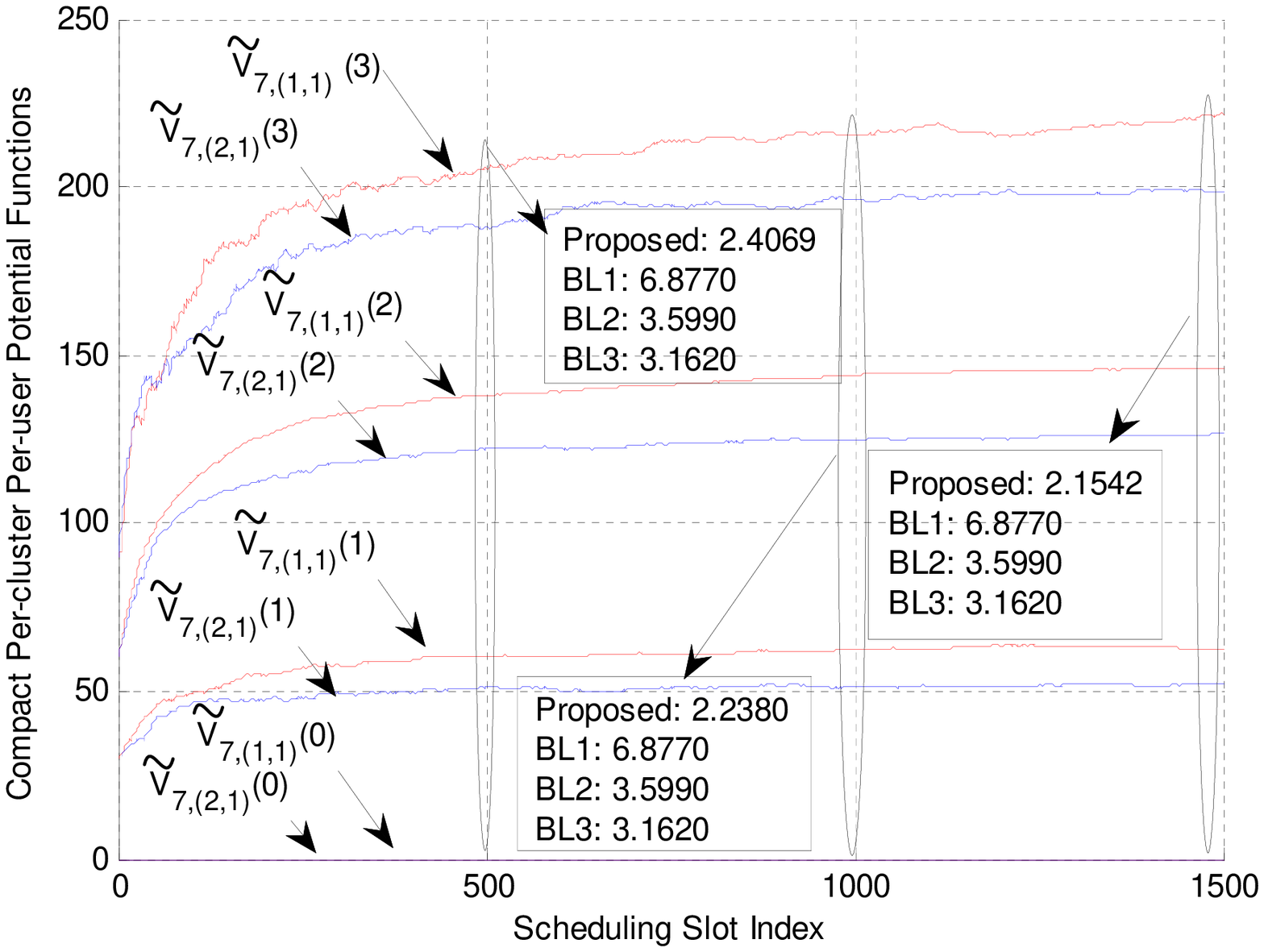}
  \caption{Convergence property of the proposed distributive
  stochastic learning algorithm via stochastic approximation. The
  transmit power $\bar P_b=35$ dbm, the maximum cluster size $N_B=3$,
  the number of transmit antenna at each BS $N_t=4$, the number of MSs
  per BS $K=1$ and the resolution level $d=3$. The figure illustrates
  instantaneous per-cluster potential function values versus
  instantaneous slot index. The boxes indicated the mean delay of
  various schemes at three selected slot indices.}
  \label{Fig:potentiallearning}
  \end{center}
  \end{figure}

Fig. \ref{Fig:potentiallearning} illustrates the convergence
property of the proposed online learning algorithm for estimating
the {\em Compact Per-cluster Per-user Potential Functions}
$\{\widetilde V^t_{n,(b,k)}\}$. We plot the transient of potential
function versus slot index at a transmit power $\overline P_b=30$ dbm. It
can be observed that the proposed distributive learning algorithm
converges quite fast. The average delay corresponding to the  the
500-th scheduling slot is 2.4069 pck, which is quite close to the
optimal delay and is much smaller than the other baselines.

\section{Summary}

In this paper, we propose a two-timescale delay-optimal dynamic
clustering and power allocation design for the downlink network MIMO
systems. We show that the delay-optimal control can be formulated as
an infinite-horizon average cost CPOMDP and derive an
\textcolor{black}{{\em equivalent Bellman equation}} to solve the
CPOMDP. To address the distributive requirement and the issue of
exponential memory requirement and computational complexity, we
propose a novel distributive online learning algorithm performed to
estimate the distributive potential functions as well as the LMs. We
also show that the proposed distributive online learning algorithm
converges almost surely (with probability 1). We formulate the
instantaneous power allocation as a {\em Per-stage QSI-aware
  Interference Game} played among all the CMs and propose a
{\em QSI-aware \textcolor{black}{Simultaneous Iterative
Water-filling Algorithm}} (QSIWFA) to achieve the NE. The proposed
algorithm achieves significant performance gains over all the
baselines due to the QSI-aware dynamic clustering and QSIWFA power
control.

\begin{appendix}

\textcolor{black}{
  \section*{Appendix A: Proof of Lemma \ref{Lem:reduced-MDP}}
Given any stationary pattern
  selection policy $\Omega_c$, by standard MDP techniques, the
  optimal power allocation policy $\Omega_p(\mathbf H, \mathbf Q)$
 can be obtained by solving the following Bellman equation:
\begin{align}
  \theta_{\Omega_c}+V_{\Omega_c}(\mathbf H,\mathbf Q)
  &=
  \min_{\Omega_p(\mathbf H,\mathbf Q)}\{g(\gamma,\mathbf H,\mathbf
  Q,\Omega_c,\Omega_p)+
  \sum_{\mathbf H',\mathbf Q'}\Pr[\mathbf H',\mathbf Q'|\mathbf H',\mathbf Q',\Omega_c,\Omega_p]V_{\Omega_c}(\mathbf H',\mathbf Q')\}\nonumber\\
  &\stackrel{(a)}{=}
  \min_{\Omega_p(\mathbf H,\mathbf Q)}\{g(\gamma,\mathbf H,\mathbf
  Q,\Omega_c,\Omega_p)+
  \sum_{\mathbf Q'}\Pr[\mathbf Q'|\mathbf H,\mathbf
  Q,\Omega_c,\Omega_p]\sum_{\mathbf H'}\Pr[\mathbf
  H']V_{\Omega_c}(\mathbf H',\mathbf Q')\}\nonumber\\
  \stackrel{(b)}{\Rightarrow} \mathbb E[\theta+ V(\mathbf H,\mathbf Q)|\mathbf Q]
  &=\mathbb E\Big[ \min_{\Omega_p(\mathbf
    H,\mathbf Q)}\{g(\gamma,\mathbf H,\mathbf Q,\Omega_c,\Omega_p)+
  \sum_{\mathbf Q'}\Pr[\mathbf Q'|\mathbf H,\mathbf
  Q,\Omega_c,\Omega_p]\sum_{\mathbf H'}\Pr[\mathbf H']V(\mathbf
  H',\mathbf Q')\Big| \mathbf Q\Big], \forall\mathbf Q
  \nonumber
\end{align}
where $\{V_{\Omega_c}(\mathbf H,\mathbf Q)\}$ are the associated
potential functions, (a) is due to (\ref{eqn:transition-prob1}) and
(b) is obtained by taking conditional expectation (average over
$\mathbf H$ conditioned on $\mathbf Q$) on both sides, due to the
i.i.d. assumption on GCSI $\mathbf H$ in Assumption \ref{Asump:H}.
The optimal $\Omega_c$ can be obtained by solving the following
Bellman equation:
\begin{align}
  \mathbb E[\theta+ V(\mathbf H,\mathbf Q)|\mathbf Q]
  =\min_{\Omega_c(\mathbf Q)}\mathbb E\Big[ \min_{\Omega_p(\mathbf
    H,\mathbf Q)}\{g(\gamma,\mathbf H,\mathbf Q,\Omega_c,\Omega_p)+
  \sum_{\mathbf Q'}\Pr[\mathbf Q'|\mathbf H,\mathbf
  Q,\Omega_c,\Omega_p]\sum_{\mathbf H'}\Pr[\mathbf H']V(\mathbf
  H',\mathbf Q')\Big| \mathbf Q\Big], \forall\mathbf Q
  \nonumber
\end{align}
where $\{ V(\mathbf H,\mathbf Q)\}$ are the associated potential
functions. Define $\overline V(\mathbf Q)=\mathbb E[V(\mathbf
H,\mathbf Q)|\mathbf Q]$, we can obtain the equivalent Bellman
equation:
\begin{align}
  \theta+\overline V(\mathbf Q)
  &=
  \min_{\Omega_c(\mathbf Q)}
  \mathbb E\Big[
  \min_{\Omega_p(\mathbf H,\mathbf Q)}\{g(\gamma,\mathbf H,\mathbf
  Q,\Omega_c,\Omega_p)+
  \sum_{\mathbf Q'}\Pr[\mathbf Q'|\mathbf H,\mathbf
  Q,\Omega_c,\Omega_p]\overline
  V(\mathbf Q')\Big|\mathbf Q\Big]\nonumber
  \\
  &\stackrel{(c)}{=}\min_{\Omega_c(\mathbf Q),\Omega_p(\mathbf Q)}
  \{\overline g(\gamma,\mathbf
  Q,\Omega_c,\Omega_p)+
  \sum_{\mathbf Q'}\Pr[\mathbf Q'|\mathbf
  Q,\Omega_c,\Omega_p]\overline V(\mathbf Q')\}, \ \ \forall\mathbf Q
\label{eqn:Bellman2}
\end{align}
where (c) is due to the definition of ``conditional action sets'' in
Definition \ref{defn:conditional-action1}.  Let
$\Omega^*=(\Omega_c^*, \Omega_p^*)$ denote the optimal control
policy minimizing R.H.S. of \eqref{eqn:Bellman2} at any state
$\mathbf{Q}$, and $\theta = L_{\beta}^*(\boldsymbol{\gamma}) =
\inf_{\Omega}L_{\beta}(\Omega,\boldsymbol{\gamma})$ is the optimal
average cost per stage. By Definition
\ref{defn:conditional-action1}, we have the associated original
control policy $\Omega^*(\boldsymbol \chi)=(\Omega_c^*(\mathbf Q),
\Omega_p^*(\boldsymbol \chi) )$, which solves Problem \ref{Prob2}
and hence, $\theta$ is also the optimal average cost per stage of
Problem \ref{Prob2}. Due to the discrete nature of pattern
selection, we introduce the {\em
  Pattern Selection Q-factor} $\mathbb Q(\mathbf
Q, C)$ ($\forall C \in \mathcal{C},\forall \mathbf Q$) to facilitate
the pattern selection, which is defined as
\begin{align}
\mathbb Q(\mathbf Q, C)\triangleq& \min_{\Omega_p(\mathbf{Q})}\Big\{
\overline{g}\big(\boldsymbol{\gamma},\mathbf{Q},
C,\Omega_p(\mathbf{Q})\big) + \sum_{\mathbf{Q}'} \Pr[\mathbf{Q}'|
\mathbf{Q}, C,\Omega_p(\mathbf{Q})] \overline{V}(\mathbf{Q}')
\Big\}-\theta
\label{eqn:pattern-selection-Q} \\
\stackrel{(d)}{=}&\mathbb{E} \Big[
A\min_{\Omega_p(\mathbf{H},\mathbf{Q})}\big\{
g\big(\boldsymbol{\gamma},(\mathbf{H},\mathbf{Q}), C,
\Omega_p(\mathbf{H},\mathbf{Q})\big) + \sum_{\mathbf{Q}'}
\Pr[\mathbf{Q}'| (\mathbf{H},\mathbf{Q}), C,
\Omega_p(\mathbf{H},\mathbf{Q})] \overline{V}(\mathbf{Q}') \big\}
\Big
| \mathbf{Q} \Big ]-\theta\nonumber\\
=& \text{R.H.S. of \eqref{eqn:Bellman3}}\nonumber
\end{align}
where (d) is due to Definition \ref{defn:conditional-action1}.
Therefore,  $\overline V(\mathbf Q)=\min_{ C}\mathbb Q(\mathbf Q,
C)$, $\Omega_c^*(\mathbf Q)=\arg \min_{ C}\mathbb Q(\mathbf Q, C)$
and $\{\mathbb Q(\mathbf Q, C)\}$ satisfies \eqref{eqn:Bellman3}. }

\section*{Appendix B: Proof of Lemma \ref{Lem:convergence-dis-fixed-gamma}}

Since the {\em per-cluster potential} function $\{\overline V_n^t(\mathbf
Q_n)\}$ of each queue state $\mathbf Q_n$ is updated comparably
often \cite{Borkarasynchronous:1998}, the only difference between
the synchronous and asynchronous update is that the resultant ODE of
the asynchronous update is a time-scaled version of the synchronous
update \cite{Borkarasynchronous:1998}, which does not affect the
convergence behavior. Therefore, we consider the convergence of
related synchronous version in the following. In the following
proof, we shall use $i$ ($1\leq i \leq I_{Q_n}$) instead of $\mathbf
Q_n^i$ for simplicity. since $\boldsymbol \gamma_n$ is quasi-static
over the timescale I, we shall omit $\boldsymbol\gamma_n$ in the
arguments of the function $\mathbf T_n(\cdot)$.


We shall first show the convergence of the martingale noise in
\eqref{eqn:per-cluster-online_learn_potential}. Let $\mathbb E_t$
and $\Pr_{t}$ denote the expectation and probability conditioned on
the $\sigma$-algebra $\mathcal F_{t}$, generated by $\{\overline{\mathbf
V}_n^0,\mathbf Y_n^k,k\le t\}$. Define $S_n^t(i)\triangleq \mathbb
E_t(Y_n^t(i)) = T_{n,i}(\mathbf {\overline V}_n^t)- {\overline V}^t(i)
-T_{n,I}(\mathbf {\overline V}_n^t)-{\overline V_n}^t(I)) $, where $Y_n^t(i)$
is the noise corrupted observation of $S_n^t(i)$ given in
\eqref{eqn:per-cluster-online_learn_potential}. Define  $\delta
M^t(i) = Y_n^t(i)-\mathbb E_{t}[Y_n^t(i)]$, which is the martingale
difference noise with property that $\mathbb E_t[\delta M_n^t(i)]=0$
and $\mathbb E[\delta M_n^t(i)\delta M_n^{t'}(i)]=0,\forall t\neq
t'$. For some $k$, define $M_n^t(i)=\sum_{l=k}^{t}\epsilon^v_l\delta
M_n^l(i)$. Thus, from \eqref{eqn:online_learn_potential}, we have
\begin{align}
\label{eqn:potential-learn-with-martingale-in-proof} \overline
V_n^{t+1}(i)=\overline V_n^{t}(i) + \epsilon^v_t [S^t(i)+\delta M_n^t(i)]
=\overline V_n^{k}(i)+ \sum_{l=k}^{t}\epsilon^v_l S^l(i) + M_n^t(i)
\end{align}
Since $\mathbb E_t[M_n^t(i)] = M_n^{t-1}(i)$, $\{M_n^t(i)\}$ is a
martingale sequence. By Kolmogorov's inequality, we have $
\Pr_k\{\sup_{k\le l\le t}|M_n^l(i)|\ge\lambda\}\le \frac{\mathbb
E_k[|\delta M_n^t(i)|^2]}{\lambda^2} \leq \overline{\delta M_n} $. By the
boundedness assumption of $\delta M_n^t(i)\ (\forall t,i)$ and the
property of the martingale difference noise as well as the condition
on the stepsize sequence in \eqref{stepsize}, we have $\mathbb
E_{k}[|M_n^t(i)|^2]=\mathbb E_{k}[|\sum_{l=k}^{t}\epsilon^v_l\delta
M_n^l(i)|^2] =\sum_{l=k}^{t}\mathbb E_{k}[(\epsilon^v_l)^2 (\delta
M_n^l(i))^2] \le \delta \overline M_n\sum_{l=k}^{t}(\epsilon^v_l)^2
\Rightarrow \lim_{k\to\infty}Pr_{k}{\sup_{k\le l \le
t}|M_n^l(i)|\ge\lambda}=0 $. Thus, as $k\to\infty$,
\eqref{eqn:potential-learn-with-martingale-in-proof} goes to $\overline
V_n^{t+1}(i)=\overline V_n^{k}(i)+ \sum_{l=k}^{t}\epsilon^v_l S_n^l(i)$
with probability 1. The vector form of update is given by:
\begin{align}
\label{eqn:learn-potential-noMG-vector-form} \mathbf {\overline
V}_n^{t+1} =  \mathbf {\overline V}_n^{k}+\sum_{l=k}^{t}
\epsilon^v_l\big[\mathbf T(\mathbf {\overline V}_n^l)-\mathbf {\overline
V}_n^l-(T_{n,I}(\mathbf {\overline V}_n^l)-{\overline V}_n^l(I))\mathbf e\big]
\end{align}

Next, we shall show the convergence of
\eqref{eqn:learn-potential-noMG-vector-form} after the Martingale
noise are averaged out. Let $\Omega_t^*$ denote the optimal control
action attaining the minimum in $\mathbf T_n(\mathbf V_n^t)$. Let
$\mathbf {\overline g}_n^t$ and $\mathbf P_n^t$ denote the conditional
per-stage reward vector and conditional average transition
probability matrix under $\Omega_t^*$. Denote $w^t = T_{n,I}(\mathbf
{\overline V}_n^{t})-{\overline V}_n^{t}(I)$, we have
\begin{align*}
S_n^t& = \mathbf{\overline g}_n^t + \mathbf P_n^t\mathbf {\overline V}_n^{t}
-\mathbf {\overline V}_n^{t} - w^t\mathbf e  \le \mathbf{\overline g}_n^{t-1}
+ \mathbf P_n^{t-1}\mathbf {\overline V}_n^{t} -\mathbf {\overline V}_n^{t} -
w^t\mathbf e
\nonumber\\
S_n^{t-1}& =  \mathbf{\overline g}_n^{t-1} + \mathbf P_n^{t-1}\mathbf
{\overline V}_n^{t-1} -\mathbf {\overline V}_n^{t-1} - w^{t-1}\mathbf e \le
 \mathbf{\overline g}_{\Omega_{t}} + \mathbf
P_n^t\mathbf {\overline V}_n^{t-1} -\mathbf {\overline V}_n^{t-1} -
w^{t-1}\mathbf e
\nonumber\\
\stackrel{\text{by iterating}}{\Rightarrow}&
\mathbf{A}_n^{t-1}\cdots\mathbf{A}_n^{t-m}S_n^{t-m} - C_1(w^t -
w^{t-m})\mathbf e \le S_n^t
\le\mathbf{B}_n^{t-1}\cdots\mathbf{B}_n^{t-m}S_n^{t-m} - C_1(w^t -
w^{t-m})\mathbf e
\end{align*}
where $C_1$ is some constant. Since $S_n^t(I)=0,\forall t$, by the
assumption in \eqref{per-cluster-promatrix-cond-dis}, we have
\begin{align*}
&(1-\delta)\min_{i'}S_n^{t-m}(i')-C_1(w^t-w^{t-m}) \le S_n^{t}(i)
\le (1-\delta)\max_{i'}
 S_n^{t-m}(i')-C_1(w^t-w^{t-m})\ \forall i
\nonumber\\
&\Rightarrow\left\{
\begin{array}{ll}
\min_{i'}S_n^t(i')\ge(1-\delta)\min_{i'}S_n^{t-m}(i')-C_1(w^t-w^{t-m})\\
\max_{i'}S_n^t(i')\le(1-\delta)\max_{i'}
S_n^{t-m}(i')-C_1(w^t-w^{t-m})
\end{array}
\right.
\end{align*}
Therefore $\max_{i'}S_n^t(i')-\min_{i'}S_n^t(i')\le
(1-\delta)\big(\max_{i'}S_n^{t-m}(i')-\min_{i'}
S_n^{t-m}(i')\big)\Rightarrow
\max_{i'}S_n^t(i')-\min_{i'}S_n^t(i')\le
\phi_k\prod_{l=1}^{\lfloor\frac{t-k}{m}\rfloor}(1-\delta_{k+lm}) $,
where $\phi_k>0$. Since $S_n^t(I)=0$, we  have $\max_{i'}
S_n^t(i')\ge0$ and $\min_{i'}S_n^t(i')\le 0$. Thus, $\forall i$, we
have $|S_n^t(i)| \le \max_{i'}S_n^t(i')-\min_{i'}S_n^t(i') \le
\phi_k\prod_{l=1}^{\lfloor\frac{t-k}{m}\rfloor}(1-\delta_{k+lm})$.
Therefore, as $t\to\infty$, $S_n^t\to 0$, i.e. $\mathbf {\overline
V}_n^{\infty}$ satisfies the fixed point equation
\eqref{bellmaneqn-wp1-dis-fixed-gamma}. $\mathbf {\overline
V}_n^{\infty}$ is the potential vector, which is up to an constant
vector\cite{Bertsekas:2007}. However, due to the property that
$S_n^t(I)=0\ \forall t\Rightarrow\overline V_n^{t}(I)=\overline V_n^{0}(I)\
\forall t$, we have the convergence of the potential vector $\mathbf
{\overline V}_n^{\infty}=\lim_{t\to\infty}\mathbf {\overline V}_n^{t}$.

\section*{Appendix C: Proof of Lemma \ref{Lem:convergence-LM}}

Due to the two time scale separation, the primal update of the {\em
per-cluster per-user potential} can be regarded as converged to
$\overline V_n^{\infty}$ w.r.t the LMs $\{\gamma_b^t\}$ at $t$-th slot.
\cite{Borkartwotimescales:1997}. Using standard stochastic
approximation theorem\cite{Borkarbook:2008}, the dynamics of the LM
$\gamma_b$ for BS $b\ (\forall b\in\mathcal B)$ update equation in
\eqref{eqn:per-cluster-online_learn_LM} can be represented by the
following ODE:
\begin{align}
 \dot{\gamma_b}(t) =
\mathbb{E}^{\big(\Omega_c^*(\boldsymbol{\gamma}(t)),\Omega_{
p}^*(\boldsymbol{\gamma}(t))\big)}[ P_b(t) - \overline P_b], \ \forall
b\in\mathcal B \label{eqn:noMG_learn_LM}
\end{align}
where $\Omega_c^*(\boldsymbol{\gamma}(t))$ is the converged policy
in  \eqref{eqn:per-cluster-approximate-Q-pattern}, $\Omega_{
p}^*(\boldsymbol{\gamma}(t))$ is the converged policy minimizing the
R.H.S. of \eqref{eqn:bellman-per-cluster-potential} for each cluster
$n$, and
$\mathbb{E}^{\big(\Omega_c^*(\boldsymbol{\gamma}(t)),\Omega_{
p}^*(\boldsymbol{\gamma}(t))\big)}[\cdot]$ denotes  the expectation
w.r.t the measure induced by
$\big(\Omega_c^*(\boldsymbol{\gamma}(t)),\Omega_{
p}^*(\boldsymbol{\gamma}(t))\big)$. Define
$G(\boldsymbol{\gamma})=\mathbb{E}^{\big(\Omega_c^*(\boldsymbol{\gamma}),\Omega_{
p}^*(\boldsymbol{\gamma})\big)}[\sum_{\omega_n\in C} \overline
g_{n}(\boldsymbol{\gamma}_n, \mathbf Q_{n}^i,\Omega_{\mathbf p_n}^*(
\mathbf Q_{n}^i))]$, where $\Omega_{ p}^*(\boldsymbol{\gamma}) =\arg
\min_{\Omega_{
p}(\boldsymbol{\gamma})}\mathbb{E}^{\big(\Omega_c^*(\boldsymbol{\gamma}),\Omega_{
p}(\boldsymbol{\gamma})\big)}[\sum_{\omega_n\in C} \overline
g_{n}(\boldsymbol{\gamma}_n, \mathbf Q_{n}^i,\Omega_{\mathbf p_n}(
\mathbf Q_{n}^i))]$. Since clustering pattern selection policy is
discrete, we have
$\Omega_c^*(\boldsymbol{\gamma})=\Omega_c^*(\boldsymbol{\gamma}+\boldsymbol{\delta}_{\gamma})$.
Hence, by chain rule, we have $\frac{\partial G}{\partial
\gamma_b}=\sum_{b',k}\frac{\partial G}{\partial
p_{b',k}}\frac{\partial  p_{b',k}}{\partial\gamma_b}+
\mathbb{E}^{\big(\Omega_c^*(\boldsymbol{\gamma}),\Omega_{
p}^*(\boldsymbol{\gamma})\big)}[ P_b(t) - \overline P_b]$. Since
$\Omega_{ p}^*(\boldsymbol{\gamma}) =\arg \min_{\Omega_{
p}(\boldsymbol{\gamma})}\mathbb{E}^{\big(\Omega_c^*(\boldsymbol{\gamma}),\Omega_{
p}(\boldsymbol{\gamma})\big)}[\sum_{\omega_n\in C} \overline
g_{n}(\boldsymbol{\gamma}_n, \mathbf Q_{n}^i,\Omega_{\mathbf p_n}(
\mathbf Q_{n}^i))]$, we have $\frac{\partial G}{\partial
\gamma_b}=0+
\mathbb{E}^{\big(\Omega_c^*(\boldsymbol{\gamma}),\Omega_{
p}^*(\boldsymbol{\gamma})\big)}[ P_b - \overline P_b]=\dot{\gamma_b}(t)$.
Therefore, we show that the ODE in \eqref{eqn:noMG_learn_LM} can be
expressed as $\dot{\boldsymbol \gamma}(t)= \triangledown
G(\boldsymbol{\gamma}(t))$. As a result, the ODE in
\eqref{eqn:noMG_learn_LM} will converge to $\triangledown
G(\boldsymbol{\gamma})=0$, which corresponds to the per-BS average
power constraints in \eqref{eqn:tx-pwr-constraint}.

\section*{Appendix D: Proof of Lemma \ref{Lem:per-cluster-per-user-conditional-potential}}

Substitute the transition probability in
\eqref{eqn:per-cluster-tran-prob} into
\eqref{eqn:bellman-per-cluster-potential} and then apply standard
optimization techniques to minimize the R.H.S. of
\eqref{eqn:bellman-per-cluster-potential}, we can obtain the
closed-form optimal power control policy for given CQSI and CCSI:
$p_{b,k}(\mathbf H_{n},\mathbf Q_n) = \Big( \frac{\frac{\tau}{\overline
N_{b,k}}\Delta_{b,k} \overline V_n(\mathbf Q_{n})}
{\sum_{b'\in\omega_n}\gamma^{b'}\parallel \mathbf{
w}_{(b,k),b'}\parallel^2} -(1+\hat I_{n,(b,k)})\Big)^+$, where
$\Delta_{b,k} \overline V_n(\mathbf Q_{n})=\overline V_n(\mathbf
Q_{n})-\overline V_n([\mathbf Q_{n}-\mathbf e_{b,k}]^+)$. Similarly,
substitute the transition probability in
\eqref{eqn:per-cluster-per-user-per-stage-cost} into
\eqref{eqn:bellman-per-cluster-per-user-potential} and then apply
standard optimization techniques to minimize the R.H.S. of
\eqref{eqn:bellman-per-cluster-potential}, we can obtain the
closed-form optimal power control policy for given LQSI and CCSI:
$p_{b,k}(\mathbf H_{n},Q) = \Big(\frac{\frac{\tau}{\overline
{N}_{b,k}}\Delta \overline V_{n,(b,k)}(Q)}
{\sum_{b'\in\omega_n}\gamma^{b'}\parallel \mathbf{
w}_{(b,k),b'}\parallel^2} -(1+\hat I_{n,(b,k)})\Big)^+$, where
$\Delta \overline V_{n,(b,k)}(Q)=\overline V_{n,(b,k)}(Q)-\overline
V_{n,(b,k)}([Q-1]^+)$.

Solution of Bellman equation in
\eqref{eqn:bellman-per-cluster-potential}  can be obtained by
offline relative value iteration\cite{Bertsekas:2007}. Without loss
of generality, we set  $\mathbf Q_n^I=\mathbf 0$ as the reference
state. Hence, we have normalizing equation $\overline V_n^l(\mathbf
Q_n^I)=0, \forall l$. Assume $\overline V_n^l(\mathbf
Q_n)=\sum_{b\in\omega_n}\sum_{k\in\mathcal K}\overline
V_{n,(b,k)}^l(Q_{b,k}), \forall l$.

At the $(l-1)$-th iteration, updating policy by minimizing the R.H.S
of \eqref{eqn:bellman-per-cluster-potential} is given by $ \hat
p_{b,k}^l(\mathbf H_{n},Q_{b,k}) = \Big(\frac{\frac{\tau}{\overline
N_{b,k}}\Delta \overline V^{l-1}_{n,(b,k)}(Q_{b,k})}
{\sum_{b'\in\omega_n}\gamma_{b'}\parallel \mathbf{
w}_{(b,k),b'}\parallel^2} -1\Big)^+$$\Rightarrow
\hat\mu_{b,k}^l(\mathbf H_n, Q_{b,k})=\mathbb E[\log(1+\hat
p_{b,k}^l) | \mathbf H_n, Q_{b,k}]/\overline N_{b,k}$, where
$\hat\mu_{b,k}^l(\mathbf H_n, Q_{b,k})$ is the mean departure rate,
and $\Delta \overline
V^{l}_{n,(b,k)}(Q_{b,k})=V^{l}_{n,(b,k)}(Q_{b,k})-V^{l}_{n,(b,k)}([Q_{b,k}-1]^+)$
is the potential increment for the MS $(b,k)$'s queue.

At the $l$-th iteration, we determine the potential $\overline
V_n^l(\mathbf Q_n)$ and $\theta_n^l$ by solving the normalizing
equation $\overline V_n^l(\mathbf Q_n^I)=0$ together with
$I_{Q_n}=(N_Q+1)^{K|\omega_n|}$ fixed point equations in
\eqref{eqn:bellman-per-cluster-potential}, which is given by
$\theta_{n,(b,k)}^l= \sum_{b\in\omega_n}\sum_{k\in \mathcal
K}{\theta_{n,(b,k)}^l}$ with \textcolor{black}{
\begin{align}
 &{\theta_{n,(b,k)}^l}=\Big( \overline
g_{n,(b,k)}(\boldsymbol{\gamma}_n,Q_{b,k},\hat p_{b,k}^l(Q_{b,k})) +
\lambda_{b,k}\tau \Delta\overline
V^{l}_{n,(b,k)}(\min\{Q_{b,k}+1,N_Q\})
-\overline{\hat\mu}_{b,k}^l(Q_{b,k})\tau \Delta\overline
V^{l}_{n,(b,k)}(Q_{b,k}) \Big) \label{eqn:decompose-poisson-eq}
\end{align}
}
where $\overline g_{n, (b,k)}$ is obtained by applying interchange order
of double summation over $b$ and $b'$ of $\overline g_n$ in
\eqref{eqn:per-cluster-per-stage-cost} and decompose it into
per-cluster per-user $\overline g_{n,(b,k)}$.
 $\overline{\hat\mu}_{b,k}^l(Q_{b,k}^i)=E[\hat\mu_{b,k}^l(\mathbf
H_{n},Q_{b,k}^i)|Q_{b,k}^i]$. There are $I_{Q_n}$ joint $\mathbf
Q_n=(Q_{(b,k)})_{b\in\omega_n,k\in\mathcal K}$ states, but there are
only $N_Q+1$ states for $Q_{b,k}\forall b\in\omega_n,k$. Hence, in
the original $I_{Q_n}$ fixed-point equations
\eqref{eqn:decompose-poisson-eq}, there are only $N_Q+1$ independent
fixed-point  equations for each MS $(b,k)$ in
\eqref{eqn:decompose-poisson-eq}. In addition, set $\overline
V_{n,(b,k)}^l(0)=0,\forall b,k$ as the individual normalizing
equation, which also satisfies $\overline V_n^l(\mathbf
Q_n^I)=\sum_{b\in\omega_n}\sum_{k}\overline V_{n,(b,k)}^l(0)=0$. Hence,
in the $l$-th iteration, we can obtain $\{\overline
V_{n,(b,k)}^l(Q_{b,k}),\theta_{n,(b,k)}^l\}$ by solving each MS's
 \textcolor{black}{equivalent Bellman  equation} in \eqref{eqn:decompose-poisson-eq}.
Accordingly, $\{\overline V_n^l(\mathbf Q_n),\theta_{n}^l\}$ is the
solution for the original $I_{Q_n}$ fixed-point  equations
\eqref{eqn:decompose-poisson-eq}, where $\overline V_n^l(\mathbf
Q_n)=\sum_{b\in\omega_n}\sum_{k}\overline V_{n,(b,k)}^l(Q_{b,k})$ and
$\theta_n^l=\sum_{b\in\omega_n}\sum_{k}\theta_{n,(b,k)}^l$.

Continue the iteration until the optimal policy converges.  We
obtain $\{\overline V_{n,(b,k)}(Q_{b,k}),\theta_{n,(b,k)}\}$, and $\{\overline
V_n(\mathbf Q_n),\theta_{n}\}$ as a solution of
\eqref{eqn:bellman-per-cluster-per-user-potential} and
\eqref{eqn:bellman-per-cluster-potential}, respectively.

\section*{Appendix E: Proof of Lemma \ref{Lem:convergence-IWFA}}

The inter-cell interference power $I_{n,(b,k)}$ is  given as $
I_{n,(b,k)}= \sum_{\substack{b'\in\omega_{n'}\\n'\neq
n}}\sum_{b''\in\omega_{n'},k''\in\mathcal K }
 \big|\mathbf{h}_{(b,k),b'}\mathbf w_{(b'',k''),b'}\big|^2
p_{b'',k''} $. since the projection of $(\cdot)^+$ is non-expansive
\cite{Bertsekasbookparalleldistributed:1989}, for any two power
allocation vector $\mathbf p_{(1)},\mathbf p_{(2)}\in\mathbb
R_+^{BK}$,
\begin{align*}
|WF_{n,(b,k)}(\mathbf p_{(1)})-WF_{n,(b,k)}(\mathbf p_{(2)})|\le
\sum_{\substack{b'\in\omega_{n'}\\n'\neq n}}
\sum_{\substack{b''\in\omega_{n'}\\k''\in\mathcal K }}
 \big|\mathbf{h}_{(b,k),b'}\mathbf w_{(b'',k''),b'}\big|^2
\cdot|p_{b'',k''(1)}-p_{b'',k''(2)}|, \ \forall b\in\mathcal B,
k\in\mathcal K
\end{align*}
put the above inequality in vector form we have
$\|\mathbf{WF}(\mathbf p_{(1)})-\mathbf{WF}(\mathbf p_{(2)})\|\leq
\|\mathbf S\cdot(\mathbf p_{(1)}-\mathbf p_{(2)})\|$, with the
matrix $\mathbf S$ defined in
\eqref{eqn:S-linear-mapping-waterfilling}. With the weighted maximum
norm $\|\cdot\|^{\mathbf u}_{\infty,\text{vec}}$ and
$\|\cdot\|^{\mathbf u}_{\infty,\text{mat}}$ defined above, $\forall
\mathbf p_{(1)},\mathbf p_{(2)}\in\mathbb R_+^{BK}$ and $\forall
\mathbf u\in\mathbb R_+^{BK}$ we have $\|\mathbf{WF}(\mathbf
p_{(1)})-\mathbf{WF}(\mathbf p_{(2)})\|^{\mathbf
u}_{\infty,\text{vec}} \le \|\mathbf S\cdot (\mathbf p_{(1)}-\mathbf
p_{(2)})\|^{\mathbf u}_{\infty,\text{vec}} \le \|\mathbf
S\|^{\mathbf u}_{\infty,\text{mat}} \|(\mathbf p_{(1)}-\mathbf
p_{(2)})\|^{\mathbf u}_{\infty,\text{vec}}$,
which is a contraction of the mapping $\mathbf{WF}$, if $\|\mathbf
S\|^{\mathbf u}_{\infty,\text{mat}}<1$ is satisfied. By the theorem
on convergence of contracting iterations
\cite{Bertsekasbookparalleldistributed:1989}, we can prove the
existence and uniqueness of NE.

\end{appendix}

\bibliographystyle{IEEEtran}
\bibliography{IEEEabrv,MC}

\textcolor{black}{
\begin{table}[h]
\centering \textcolor{black}{
    \begin{tabular}{|l|l|}
      \hline
      Scheme & CPU Time (s)\\
      \hline
      BL1: FCA & 0.2218e-004\\
      \hline
      BL2: Static Clustering (on CSI) & 1.9098e-004\\
      \hline
      BL3: Greedy Dynamic Clustering (on CSI) & 0.0012\\
      \hline
      Proposed Queue-aware Dynamic Clustering & 0.0094\\
      \hline
    \end{tabular}
} \caption{\textcolor{black}{{\em Running time complexity of the
baselines and the proposed scheme. The number of cells $B=19$, the
maximum cluster size $N_B=3$, the number of MSs per BS $K=1$, the
number of antenna per-BS $N_t=4$, the average arrival rate
$\lambda_{b,k}=10$ pck/slot and the resolution level
  $d=3$.}}}\label{tab:complexity}
  \end{table}
}

  %

\end{document}